%% file: neurips_data_2022.tex
\documentclass{article}

\PassOptionsToPackage{numbers}{natbib}
\usepackage[final]{neurips_data_2022}

\usepackage[utf8]{inputenc} % allow utf-8 input
\usepackage[T1]{fontenc}    % use 8-bit T1 fonts
\usepackage[pagebackref]{hyperref}       % hyperlinks
\usepackage{url}            % simple URL typesetting
\usepackage{booktabs}       % professional-quality tables
\usepackage{amsfonts}       % blackboard math symbols
\usepackage{nicefrac}       % compact symbols for 1/2, etc.
\usepackage{microtype}      % microtypography
\usepackage{xcolor}         % colors
\usepackage{latexsym}
\usepackage{comment}
\usepackage{amsmath,amssymb}
\usepackage{microtype}
\usepackage{color}
\usepackage{multicol}
\usepackage{subcaption}
\usepackage{xcolor,colortbl}

\usepackage{multirow}
\usepackage{subcaption}
\usepackage{graphicx}
\usepackage{comment}
\usepackage[export]{adjustbox}

\usepackage[normalem]{ulem}

\usepackage{wrapfig}

\definecolor{mygreen}{HTML}{008000}

\newcommand{\ouracronym}[0]{WinoGAViL}
\newcommand{\ourwebsite}[0]{\url{https://winogavil.github.io/}}
\newcommand{\associationinstance}[0]{\emph{association instance}}

\newcommand{\swowsplit}[0]{\emph{\textit{SWOW} vision baseline dataset}}
\newcommand{\gamesplit}[0]{\emph{\textit{\ouracronym{}} dataset}}

\title{\ouracronym{}:
Gamified Association Benchmark \\ to Challenge Vision-and-Language Models}

\author{
  Yonatan Bitton$^{\dagger}$\thanks{Equal contribution.} \;\; 
 Nitzan Bitton-Guetta$^{\ddagger}$\footnotemark[1] \;\; 
 Ron Yosef $^{\dagger}$ \;\; 
 Yuval Elovici$^{\ddagger}$ \;\; 
 \\ 
  \textbf{Mohit Bansal}$^{\P}$ \;\;
 \textbf{Gabriel Stanovsky}$^{\dagger}$ \;\;
 \textbf{Roy Schwartz}$^{\dagger}$\\
 $^{\dagger}$The Hebrew University of Jerusalem \quad
 $^{\ddagger}$Ben Gurion University \\
  $^{\P}$University of North Carolina at Chapel Hill \\
  \{nitzangu,elovici\}@bgu.ac.il; mbansal@cs.unc.edu \\
    \{yonatan.bitton,ron.yosef,gabriel.stanovsky,roy.schwartz1\}@mail.huji.ac.il \\
}

\begin{document}

\maketitle

\setcounter{footnote}{0} 
\begin{abstract}
  \input{sections/01_abstract}
\end{abstract}

\section{Introduction}

\input{sections/02_introduction}

\section{The \ouracronym{} Benchmark}
\input{sections/04_the_gvlab_dataset}

\section{Experiments}
\label{sec:experiments}
\input{sections/05_experiments}

\section{Related Work}
\input{sections/03_related_work}

\section{Limitations and Conclusions}
\input{sections/06_limitations_and_conclusions}

\section*{Acknowledgements}
\input{sections/acknowledgement}

\bibliographystyle{unsrtnat}
\bibliography{ref}

\section*{Checklist}
\input{sections/paper_checklist}
\newpage
\appendix
\section{Appendix}
\input{sections/99_appendix}

\end{document}

%% file: sections/01_abstract.tex
While vision-and-language models perform well on tasks such as visual question answering, they struggle when it comes to basic human commonsense reasoning skills. In this work, we introduce WinoGAViL: an online game of vision-and-language associations (e.g., between \emph{werewolves} and \emph{a full moon}), used as a dynamic evaluation benchmark. Inspired by the popular card game Codenames, a ``spymaster’' gives a textual cue related to several visual candidates, and another player tries to identify them. Human players are rewarded for creating associations that are challenging for a rival AI model but still solvable by other human players. We use the game to collect 3.5K instances, finding that they are intuitive for humans ($>$90\% Jaccard index) but challenging for state-of-the-art AI models, where the best model (ViLT) achieves a score of 52\%, succeeding mostly where the cue is visually salient. Our analysis as well as the feedback we collect from players indicate that the collected associations require diverse reasoning skills, including general knowledge, common sense, abstraction, and more. We release the dataset, the code and the interactive game, allowing future data collection that can be used to develop models with better association abilities.\footnote{\href{https://winogavil.github.io/}{\textbf{\textcolor{purple}{https://winogavil.github.io/}}}}

%% file: sections/02_introduction.tex
\input{figs_and_tables/fig1}

Humans can intuitively reason about how a cue is associated with an image \cite{de2018visual,de2021visual,liuzzi2017explicit}. For example, in Figure~\ref{fig:fig1}, the word \textit{werewolf} may be intuitively associated with images of a puppy and a full moon. These reasoning skills go beyond object detection and similarity and require rich cultural and world knowledge. 
Cognitive studies suggest that this kind of associative thinking involves connecting distant concepts in the human memory, organized as a network of interconnected ideas \cite{ovando2021brain,beaty2021forward,levy2021unveiling,de2017large,wulff2019new}. On the other hand, vision-and-language models often fail when faced with tasks that require commonsense reasoning and cultural knowledge \cite{zellers2019recognition,thrush2022winoground,hessel2022androids,talmor2022commonsenseqa}, motivating the construction of a challenging high quality vision-and-language benchmark.

In this work, we introduce a \textbf{G}amified \textbf{A}ssociation benchmark to challenge \textbf{Vi}sion-and-\textbf{L}anguage models (WinoGAViL). Inspired by Winograd Schema Challenge \cite{levesque2012winograd}, we suggest \ouracronym{} as a benchmark for multimodal machine commonsense reasoning and association abilities.
Similar to the Codenames game,\footnote{\url{https://en.wikipedia.org/wiki/Codenames_(board_game)}} each instance in \ouracronym{} is composed of a textual cue, a number $k$, and a set of candidate images. The task is to select the $k$ images most associated with the cue. We refer to the cue and the associated images as an \associationinstance{}. For example, in Figure~\ref{fig:fig1}, the pictures of a \emph{puppy} and a \emph{moon} are (arguably) the ones most associated with the cue \emph{werewolf} out of the given candidates.

\input{figs_and_tables/fig2_examples}

We develop an online game to collect novel and challenging associations. The game is used to collect data for this work, but more importantly---to serve as a dynamic source for additional data in the future. As exemplified in Figure~\ref{fig:fig1}, a ``spymaster'' first composes a new association cue given a set of images. A rival AI model (CLIP RN50 \cite{radford2021learning}) then predicts the given association, and the spymaster is rewarded inversely to its performance, motivating the spymaster to make the cue challenging. Lastly, three human players attempt to solve the association task. The spymaster is rewarded according to their performance, motivating the spymaster to compose associations that are solvable by humans and, thus, ideally more natural than examples designed to fool a model. We use crowdworkers to collect 3.5K test instances. See Figure~\ref{fig:fig2} for a collected example. 

We evaluate several state-of-the-art models on \ouracronym{} data. We find that our game allows the collection of associations that are easy for humans ($>$90\% Jaccard index) and challenging for models ($\sim$52\%), even those that are orders of magnitude larger than the model used to create the game. Our analysis shows that models succeed mostly where the cue is visually salient. Finally, we compare our collected data with data we collected via an alternative data generation baseline that relies on SWOW \cite{de2019small}, a hand-crafted resource of textual associations. Our results show that while the two approaches are relatively easy for humans, data generated by \ouracronym{} is much more challenging to machines, highlighting the value of our gamified data collection framework.

%% file: figs_and_tables/fig1.tex
\begin{figure}[!tb]
\centering
\newcommand{\figlen}[0]{\columnwidth}
    \includegraphics[width=0.8\figlen]{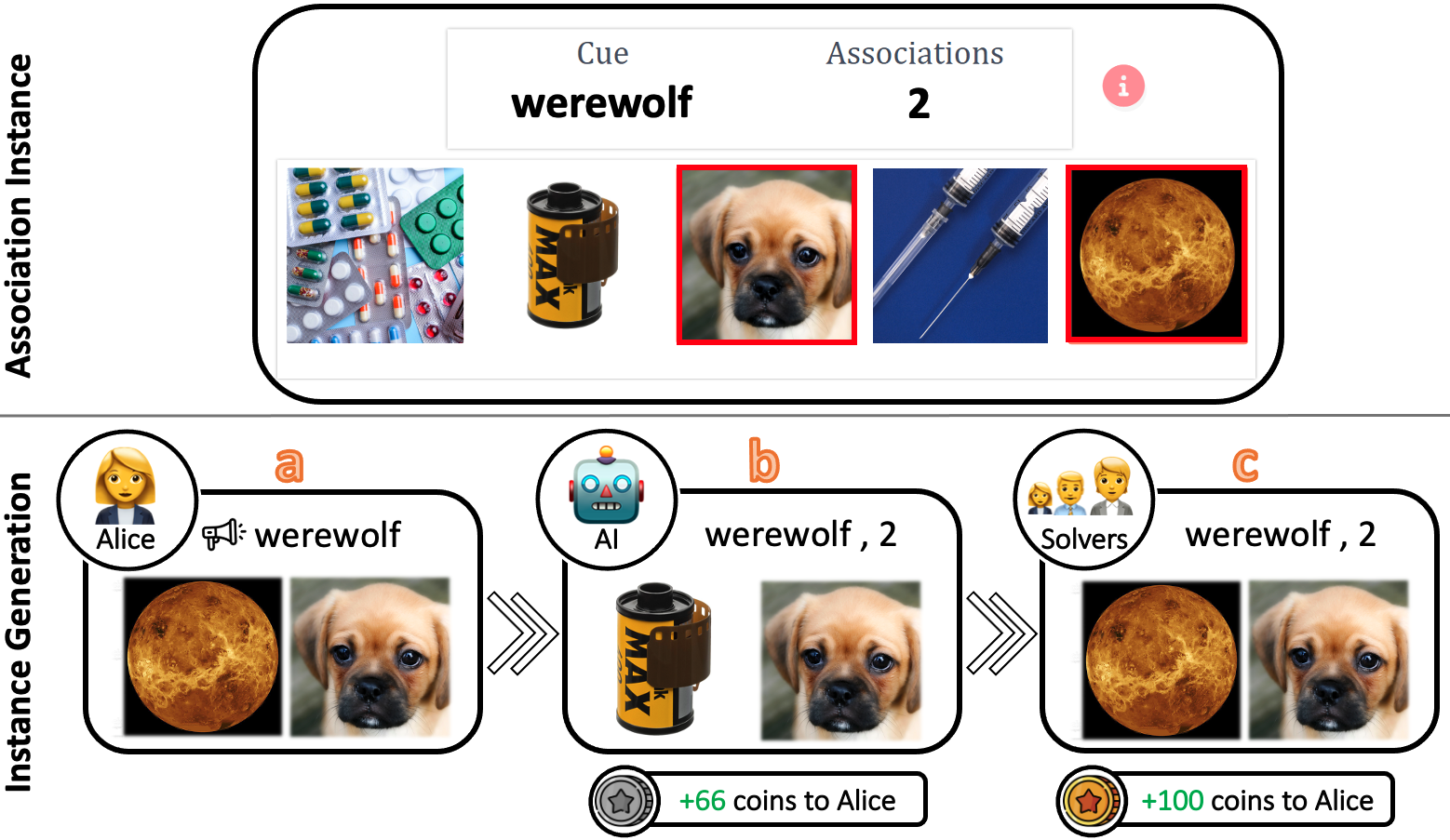}\\
    \caption{Top: An association instance from the \ouracronym{} benchmark. The task is to choose the top $k$ images that suit the cue word. In this example, the top $k$=2 images that suit the cue \textit{werewolf} are surrounded by red bounding boxes. Bottom: Game Setup---a new association instance generation. A spymaster (Alice) composes a new association given a set of images that is challenging for the rival AI model but easy for other human players. (a) Alice generates a cue word for a subset of the images; (b) A rival AI model makes a prediction based on the given cue, and Alice is rewarded inversely to the model performance; (c) Three human solvers also try to solve the task and the spymaster is rewarded according to their performance.
    }
    \label{fig:fig1}
\end{figure}

%% file: figs_and_tables/fig2_examples.tex
\begin{wrapfigure}{R}{0.5\textwidth}
%   \resizebox{1\columnwidth}{!}{\includegraphics[width=\linewidth]{images/fig2.png}
  \includegraphics[width=\linewidth]{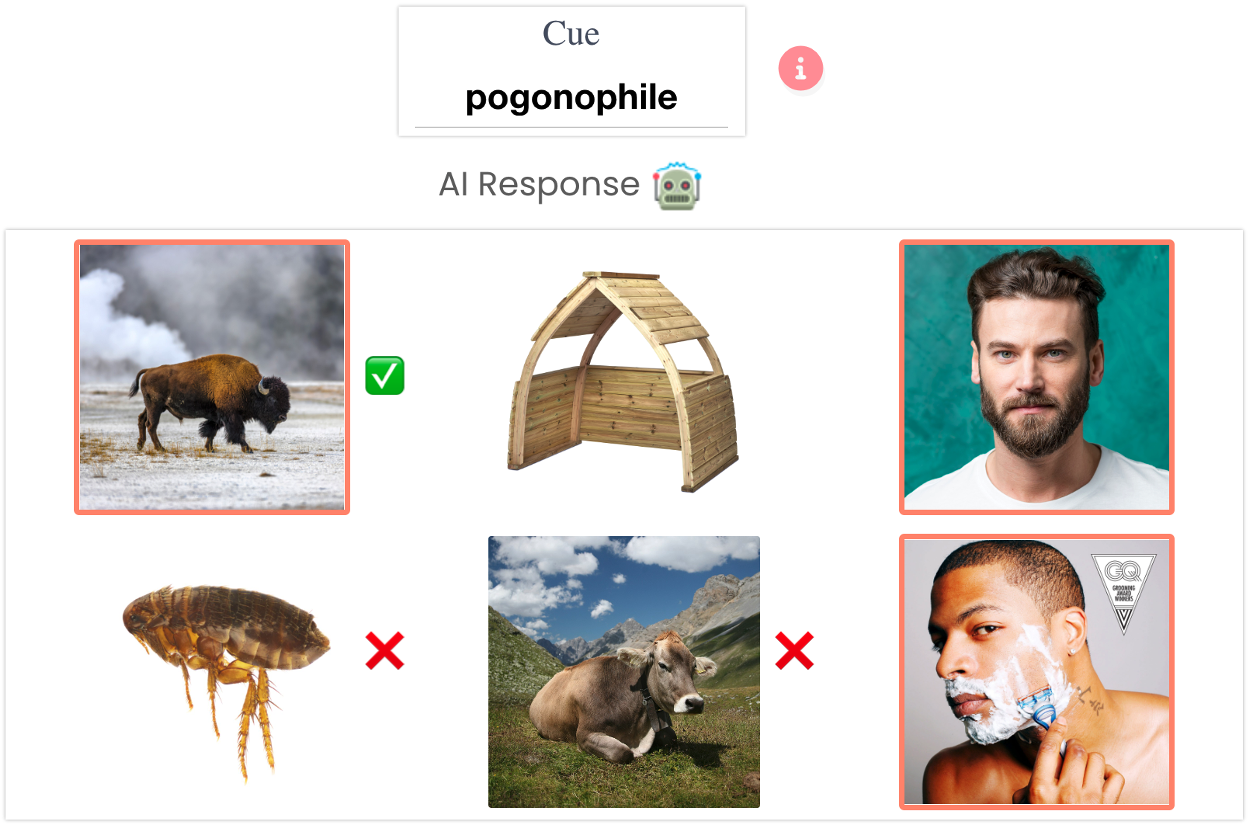}
    \caption{
      The spymaster screen for an example collected via the \ouracronym{} benchmark. The spymaster submitted the cue `pogonophile' (a lover of beards), and associated it with the three images surrounded by red bounding boxes. Model predictions are marked with V for success and X for failure. In this example the spymaster has managed to partially fool the AI model, while three other humans are able to solve it perfectly.}
  \label{fig:fig2}
\end{wrapfigure}

%% file: sections/04_the_gvlab_dataset.tex
We start by presenting the game as a framework for collecting challenging associations~(\S\ref{sec:gvlab_game}). Second, we describe how we crowd-source a test set using the game ~(\S\ref{sec:human_evaluation_and_annotation}). Finally, we analyze the collected dataset and provide statistics~(\S\ref{sec:gvlab_statistics}).

\input{sections/04A_extracting_images}

\subsection{The Game}
\label{sec:gvlab_game}
\input{sections/04B_gvlab_game}

\subsection{Human Annotation}
\label{sec:human_evaluation_and_annotation}
\input{sections/04C_human_annotation}

\subsection{\ouracronym{} Analysis}
\label{sec:gvlab_statistics}
\input{sections/04D_gvlab_statistics}

%% file: sections/04A_extracting_images.tex
\label{sec:metric_for_success}
Throughout this paper we use the Jaccard index, which is the intersection of selected candidates divided by the union of selected candidates.\footnote{\url{https://en.wikipedia.org/wiki/Jaccard_index}.} This metric does not reward random guesses highly. The random expected Jaccard index is 38\%, 34\%, 24\%, 17\% with 5/6/10/12 candidates respectively. For example, in Figure~\ref{fig:fig1}c the Jaccard index (`Human score') of the solvers is 100\%, since the intersection of the selections is the same as the union. In Figure~\ref{fig:fig1}b the AI model selection is 1/3, so the Jaccard index (`Model score') is 33\%: there are three images in the union, and one image in the intersection. 

%% file: sections/04B_gvlab_game.tex
This section describes the \ouracronym{} game environment. Besides collecting the data presented in this paper, the game can also serve as a dynamic source of new data in the future. The game setup is described below in sequential order.

% \begin{enumerate}[leftmargin=*]
\begin{enumerate}
    \item \textbf{A spymaster creates a challenging association.} A spymaster composes a new \associationinstance{} given a random set of images sampled from the web (see details below). We experiment with sets of 5, 6, 10 or 12 images. The spymaster then submits a single-word cue and selects the subset of associated images. The goal is for the association to be solvable by humans but not by the AI model. For example in Figure~\ref{fig:fig1}, the spymaster composes the cue \emph{werewolf} and associates it with the images of the \emph{puppy} and the \emph{moon}.
    \item \textbf{A rival AI model makes a prediction.}
    We then feed the association instance to a rival AI model, and report the model score. For example, in Figure~\ref{fig:fig2}, the model predicts correctly one candidate (the image of the \textit{bison}), and the total number of candidates involved is 5 :the three images the user selected and the two images falsely predicted by the model. Therefore, the model's Jaccard index is 1/5=20\%. The spymaster is rewarded inversely to the model performance, so their ``fool-the-AI'' score is (100 - `model score`) = 80\%. 

    \item \textbf{Three human players validate the created association.}
    We then give the association to three human validators, who are rewarded according to their Jaccard index for solving the association. Importantly, the spymaster's association ``solvable-by-humans'' score is determined by the average score of the three solvers. For example, in Figure~\ref{fig:fig1} all players solve the created associations perfectly; therefore, the spymaster's association ``solvable-by-humans'' score is 100\%. 
\end{enumerate}

Each player alternates between spymaster and solver roles. Each new association instance created by the spymaster is assigned to three solvers. Once the spymaster creates an association instance, their role changes to a solver responsible for solving other players' associations. This balanced approach ensures that all new associations are automatically validated by three other players. 

\paragraph{Rival AI model.} We use CLIP \cite{radford2021learning}, with a textual prompt of ``A/An [cue]''. We intentionally use a small version of CLIP (RN50), so we could evaluate the generated data with larger models. Our experiments (\S\ref{sec:experiments}) show that this data is indeed challenging for orders-of-magnitude larger models. Future versions of the game will use newer and stronger models, that are likely to further improve the data quality.

\paragraph{Image extraction.} \phantomsection
\label{sec:image_extraction}
We start with a corpus of English concepts obtained from SWOW \cite{de2019small}.\footnote{We removed words that are potentially offensive using \url{https://pypi.org/project/profanity-filter/}.} We collect an image for each concept from Google Images Download. We filter images of written words using an OCR model \cite{baek2019character}. We removed images containing the query text using an OCR model (e.g., OCR prediction ``brary'' for search query ``library''). We extract the top image based on google ranking ($\sim$2\% of the images are filtered). We also manually filter and verify that there are no inappropriate images. The result is a set of 3K images. 

\paragraph{\ouracronym{} game properties.} \ouracronym{}'s main goal is to serve as a dynamic benchmark that remains relevant as the field advances. To achieve this, we publicly release the \ouracronym{} web game, allowing dynamic data collection. The players who create associations observe the AI model predictions in real-time. Players switch roles, validating each created association as part of the game. We use rewards to motivate players to create high-quality data according to our metrics. Players are rewarded for both fooling the AI model and making the associations solvable by other humans, preventing the data from becoming unnatural and biased towards only fooling the AI model. The publicly released game includes a player dashboard and a leaderboard. All of these aim to motivate the players to compete with the AI model and with each other, leading to enhanced user engagement and high-quality data.

%% file: sections/04C_human_annotation.tex
\input{figs_and_tables/fig_mturk_ui_spymaster_example}
We hire Amazon Mechanical Turk workers to play the \ouracronym{} game. We develop qualification tests to select high-quality annotators and collect the annotators' demographic information. Spymasters screen example is presented in Figure~\ref{fig:mturk_ui_spymaster}; See Appendix~\ref{sec:appendix} for more details.\footnote{We note associations can be subjective and culture-dependent. In Section~\ref{sec:experiments} we show high agreement between our annotators.} We have several options for the total number of candidates: 5, 6, 10 or 12. With more candidates, the task naturally becomes harder. The spymasters are allowed to select between 2-5 images. Full annotation results and statistics are presented in Table~\ref{tab:annotation_statistics}. The scores of both humans and models is the Jaccard index of between their created associations instances. The annotation task includes three steps, elaborated below. 
\input{figs_and_tables/table_mturk_annotation}

First, we create new associations by asking three spymasters to create two different cues and associated candidates for a given set of images. The created association should fool the AI model but still be solvable by other humans. To reinforce it, the spymasters receive a bonus payment if their ``solvable-by-humans'' score is at least 80\%, which grows according to their ``fool-the-AI'' score, see full details of the bonus in Appendix~\ref{sec:appendix}, Section~\ref{sec:bonus}. The first row in Table~\ref{tab:annotation_statistics} presents the number of generated associations, and the second row presents the average model score (or 100-``fool-the-AI score''). The low model scores indicate that the spymasters succeeded in creating data that fools the AI model.

Second, we take the associations created via the game and ask three annotators to solve them. We compute an average Jaccard index of the three solvers for each instance. The third row in Table~\ref{tab:annotation_statistics} presents the average human score (or the spymaster's ``solvable-by-humans'' score), indicating that the spymasters were able to create data that is solvable by other humans.

Finally, we select the \ouracronym{} test set. To obtain the final test instances, we select associations solved with a mean Jaccard index of at least 80\%. The threshold can be lowered to receive more data of lower quality or raised to receive less data of higher quality. Note that in order to reduce the dependence on a specific model, we do not use the model scores in the data selection, i.e., instances that can be solved by the AI model are not automatically excluded, only the solvable-by-humans score is considered in the discarding decision. The last row in Table~\ref{tab:annotation_statistics} presents the final number of instances accumulated in the dataset.

The annotators were paid an average of 14 USD per hour for the annotation tasks (including bonuses). The total project annotation budget was 2,000 USD. The annotators received daily feedback on their performances, scores, and the bonuses they won. We denote the data created by the \ouracronym{} game by \gamesplit{}. 
In \S\ref{sec:experiments} we show that this data is easy for humans and challenging for state-of-the-art models.

%% file: figs_and_tables/fig_mturk_ui_spymaster_example.tex
\begin{figure}[!t]
\centering
   \includegraphics[width=\linewidth]{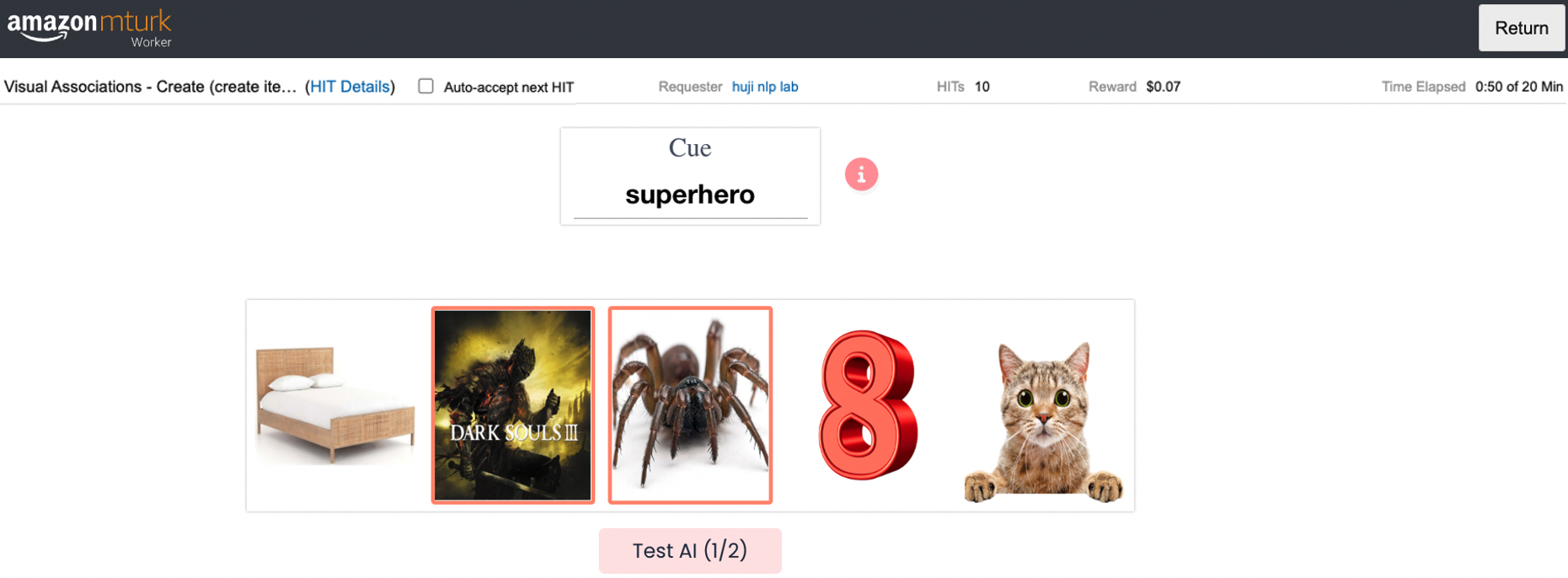}
   \caption{A screenshot from a spymaster screen in Amazon Mechanical Turk.}
\label{fig:mturk_ui_spymaster}
\end{figure}
% \FloatBarrier

%% file: figs_and_tables/table_mturk_annotation.tex
\begin{wraptable}{R}{0.53\textwidth}
\centering
\caption{\ouracronym{} collection statistics. Small differences exist between 5 and 6 candidates, and between 10 and 12 candidates, so we analyze these groups together. Compared to humans, the model struggles with increased number of candidates. }
\begin{tabular}{lll} \toprule
\# Candidates                      & 5 \& 6 & 10 \& 12 \\ \midrule
\# Generated Associations & 4,482        & 1,500    \\
\% Avg. Model Score & 50\%       & \textbf{35}\%     \\
\% Avg. Human Score & 84\%        & \textbf{80}\%    \\
\# $\geq$80\% Avg. Human Score  & 2,714      & 854  \\
\bottomrule
\end{tabular}
\label{tab:annotation_statistics}
\end{wraptable}

%% file: sections/04D_gvlab_statistics.tex
\label{sec:gvlab_analysis}
\input{figs_and_tables/table_reasoning_skills_short} 

\paragraph{Reasoning skills.}
\label{sec:reasoning_skills} We analyze the different skills required to solve the \gamesplit{}. We randomly sample 320 samples of \gamesplit{} and manually annotate these skills, observing the patterns required for humans to solve each association. Table~\ref{tab:reasoning_skills_short} presents some of the observed patterns, required skills, and frequencies. Appendix~\ref{sec:appendix}, Table~\ref{tab:reasoning_skills} presents the full table and Figure~\ref{fig:examples_visual_reasoning} presents examples of the visual associations. We see that solving \gamesplit{} requires diverse commonsense skills.

\paragraph{Players feedback.}
We collected qualitative and quantitative feedback from the crowdworkers. Table~\ref{table:players_feedback} presents quantitative questions and ratings, showing our game is recommended as an online game, is fun and has an intuitive user interface. We also asked the spymasters open questions about how seeing the AI model prediction and the performance bonus affected them. They mostly responded that these decisions were effective---\textit{``I used the model's guesses to make my associations better. I went after associations that the model frequently got wrong.''} and \textit{``bonus keep motivation up when it was hard to come up with connections''}. Full qualitative responses (open text) are presented in Section~\ref{sec:players_feedback} at Appendix~\ref{sec:appendix}.

Section~\ref{sec:additional_analysis} in Appendix~\ref{sec:appendix} includes additional analysis, for example annotator statistics with demographic information and average performance, and generated cues statistics including richness ratings of the created cues, ratings for abstract and concrete cues, and more.

\input{figs_and_tables/table_players_feedback}

%% file: figs_and_tables/table_reasoning_skills_short.tex
\begin{table}[!tb]
\caption{Some of the skills and observed patterns required to solve \ouracronym{} associations. Each association instance may require multiple skills.}
\label{tab:reasoning_skills_short}
\resizebox{\textwidth}{!}{
\begin{tabular}{@{}lllll@{}} \toprule
\textbf{Skill}&{\begin{tabular}[c]{@{}l@{}}\textbf{Observed}\\\textbf{Pattern}\end{tabular}}                                & \textbf{Description}                                                                                                    & \textbf{Example} & \textbf{\%}                                                                                                                                              \\ \midrule
\multirow{6}{*}{\begin{tabular}[c]{@{}l@{}}Non-\\Visual\end{tabular}}
 &{Attribute}           & \textcolor{red}{Cue} has attributes of \textcolor{mygreen}{Association}                                                                              & \textcolor{red}{iguana} has \textcolor{mygreen}{green} color& {14\%}                                                                                                                               \\
                                     && \textcolor{red}{Cue} is \textcolor{mygreen}{Association}                                                                                             & \textcolor{red}{miners} are \textcolor{mygreen}{dirty} &                                                                                                                                     \\ \cmidrule(l){2-5}

& {Use-Of}              & \textcolor{red}{Cue} uses the \textcolor{mygreen}{Association}                                                                                       & \textcolor{red}{miner} uses \textcolor{mygreen}{tractor}   & {9\%}                                                                                                                                  \\
                                     && \textcolor{mygreen}{Association} is used in relation to \textcolor{red}{Cue}                                                                         & \textcolor{red}{tupperware} is used to store \textcolor{mygreen}{food} &                                                                                                                     \\ \cmidrule(l){2-5} 
&\multirow{2}{*}{\begin{tabular}[c]{@{}l@{}}General\\Knowledge\end{tabular}} & \textcolor{red}{Cue} is a name for \textcolor{mygreen}{Association}                                                                & \textcolor{red}{ford} is a name of a \textcolor{mygreen}{car}  & {13\%}                                                                                                                   \\
 & & \textcolor{mygreen}{Association} is used in a relation to \textcolor{red}{Cue}                                                                & \textcolor{mygreen}{Oats} improve \textcolor{red}{horses} performance  &                                                                                                                  \\
 \midrule   
                          
\multirow{4}{*}{Visual}&{Activity}  
                                     & \multirow{1}{*}{\textcolor{mygreen}{Association}s perform a \textcolor{red}{Cue}} in the image                                                             &  {\begin{tabular}[c]{@{}l@{}}\textcolor{mygreen}{deer} \& \textcolor{mygreen}{snowman} looks like they \textcolor{red}{stare}\\ (Figure \ref{fig:examples_visual_reasoning}b)\end{tabular}} &  {6\%}                                                                                                  \\ \cmidrule(l){2-5}
                                    
&{Analogy}             & \begin{tabular}[c]{@{}l@{}}\textcolor{red}{Cue} can look like \textcolor{mygreen}{Association},\\
despite being from different concept maps \end{tabular}                                                                             & {\begin{tabular}[c]{@{}l@{}}\textcolor{mygreen}{deer} \& \textcolor{mygreen}{TV antenna} looks like a \textcolor{red}{horn}\\ (Figure \ref{fig:examples_visual_reasoning}d)\end{tabular}}     & {4\%}                                                                                                                                                                                                                     \\\cmidrule(l){2-5}
&{\begin{tabular}[c]{@{}l@{}}Visual\\Similarit\end{tabular}}             & \textcolor{mygreen}{Association} is visually similar to the \textcolor{red}{Cue}                                                                               & {\begin{tabular}[c]{@{}l@{}}The \textcolor{mygreen}{sponge} shape is similar to a \textcolor{red}{box}\\ (Figure \ref{fig:examples_visual_reasoning}a)\end{tabular}}     & {20\%}                                                                                                                                                                                                                                   \\
 \bottomrule  
\end{tabular}
}
\end{table}

%% file: figs_and_tables/table_players_feedback.tex
\begin{table}[!t]
\caption{Players feedback collected from the crowdworkers players (scale of 1-5)}
\resizebox{\textwidth}{!}{
\begin{tabular}{@{}lcccccc@{}}
\toprule
\multicolumn{7}{c}{Rate for the following skills how much you found them required while performing the task}                                     \\ \midrule
Role & Visual Reasoning & General Knowledge & Associative Thinking & Commonsense & Abstraction & Divergent Thinking  \\ \midrule
Spymaster & 4.4              & 3.6               & 4.5                  & 3.9         & 4.3                          & 4.5                              \\
Solver & 4.4              & 4                 & 4.7                  & 4.3         & 4.1                            & 4.1                               \\ 
\end{tabular}}
\resizebox{\textwidth}{!}{
\begin{tabular}{@{}lccc@{}}
\toprule
Role & Interest in play and recommend it as an online game & Level of enjoyment while doing the task & How clear was the UI \\ \midrule
Spymaster & 3.8                                                       & 3.7                                          & 4.7                                           \\
Solver & 4.1                                                       & 4.4                                          & 4.9                                           \\ \bottomrule
\end{tabular}
}
\label{table:players_feedback}
\end{table}

%% file: sections/05_experiments.tex
In this section, we provide an extensive evaluation of \gamesplit{}. First, we show the value of our game, by comparing it to an alternative data generation baseline based on SWOW \cite{thawani2019swow}, an existing resource of textual associations. We then evaluate human and models performance on both datasets and provide analysis. 
\input{sections/05A_extracting_swow_dataset}

\subsection{Evaluation Setup}
We experiment with state-of-the-art models and compare them to humans on the \gamesplit{} and the \swowsplit{}. On the \gamesplit{} we compare cases with 5-6 candidates and cases with 10-12 candidates. We use the Jaccard index as an evaluation metric (\S\ref{sec:metric_for_success}).
\input{figs_and_tables/fig_swow_explaination}
\input{sections/05B_gvlab_humans_evaluation}

\input{sections/05C_gvlab_model_evaluation}

\input{figs_and_tables/table_swow_vs_game}
\newpage
\subsection{Results and Model Analysis}
\input{sections/05D_gvlab_results}

\input{sections/05E_model_strength_and_weaknesses}

\input{sections/05F_image2text}

%% file: sections/05A_extracting_swow_dataset.tex
\subsection{Extracting the SWOW Baseline Dataset}\label{sec:swow}
We describe an alternative data generation baseline based on the SWOW (``Small World of Words'') dataset.\footnote{licensed under a Creative Commons Attribution-NonCommercial-NoDerivs 3.0 Unported License.} SWOW is an ongoing project where participants are presented with a cue word and asked to respond with the first three words that come to mind. We use a common representation of SWOW as a graph network.\footnote{https://smallworldofwords.org/en/project/explore} We select random distractors that are not associated with the cue in the SWOW graph. We combine the distractors to the association instances from SWOW and create 1,200 multiple-choice instances with 5 or 6 candidates. Each concept's image is obtained from the extracted images (§\ref{sec:gvlab_game}). Note that SWOW is based on textual associations, which were provided by humans given a cue, making it textual and non-adversarial, whereas WinoGAViL is based on visual and adversarial associations, where humans create a new cue given a set of images. Figure~\ref{fig:swow_vs_winogavil} illustrates this difference. As we did in the \ouracronym{} game, we validate with human annotation and only keep instances with a mean Jaccard score of at least 80\%. Human performance is 85\%, so most association instances are retained. The final dataset, denoted \swowsplit{}, is composed of 1,000 instances.

%% file: figs_and_tables/fig_swow_explaination.tex
\begin{figure}[!t]
\centering
\begin{subfigure}{.48\textwidth}
  \centering
  \includegraphics[width=\linewidth]{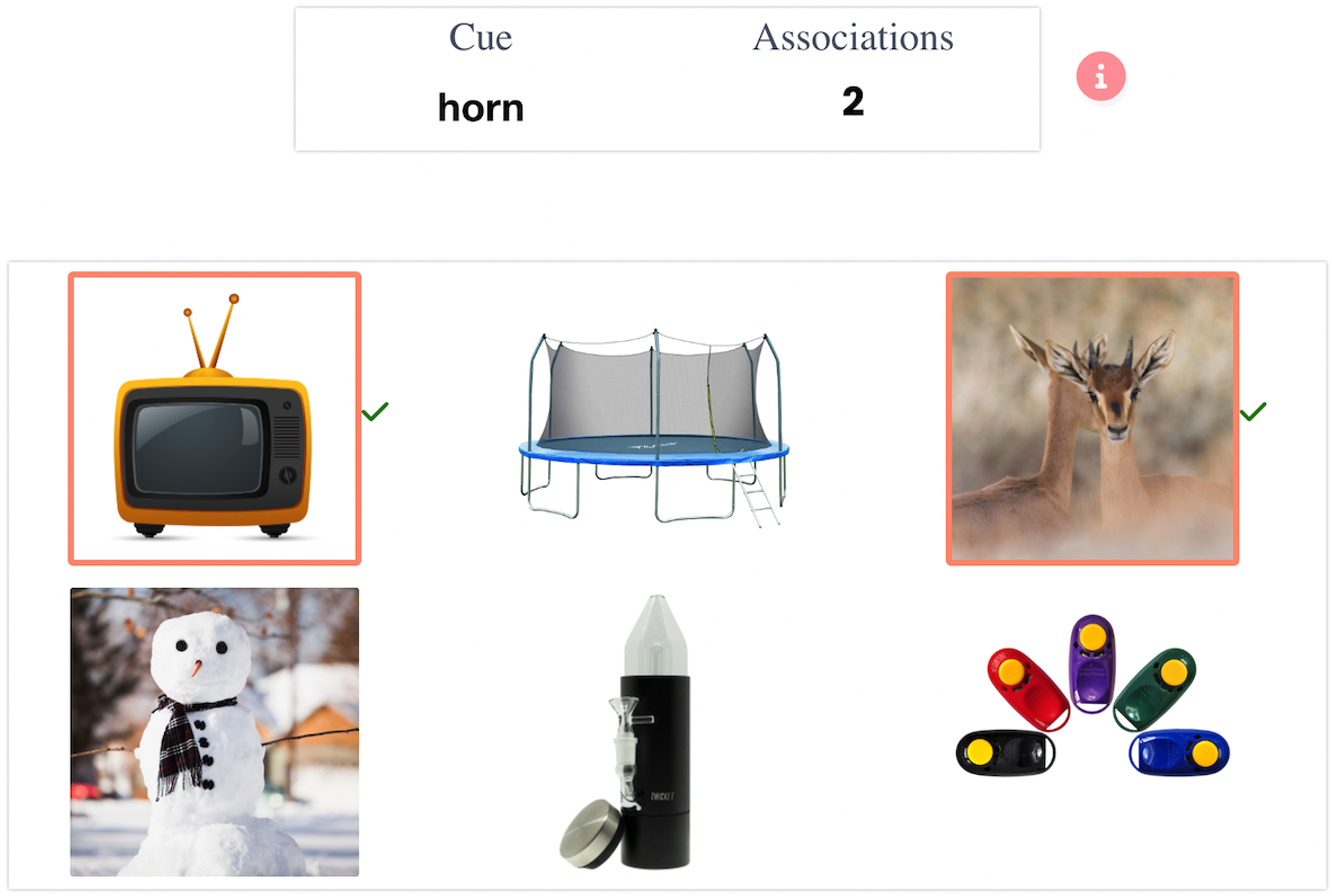}
  \caption{An association from \gamesplit{}, collected by our interactive game. Spymaster composes a new association given a set of images, aiming to fool a rival AI model. The spymaster has created the cue \emph{horn} and selected the two images surrounded by bounding boxes. This association instance cannot be solved without the specific image information (TVs usually don't have horns). Cues are assigned in a visual and adversarial manner.}
\end{subfigure}%
\hspace{2mm}
\begin{subfigure}{.48\textwidth}
  \centering
  \includegraphics[width=\linewidth]{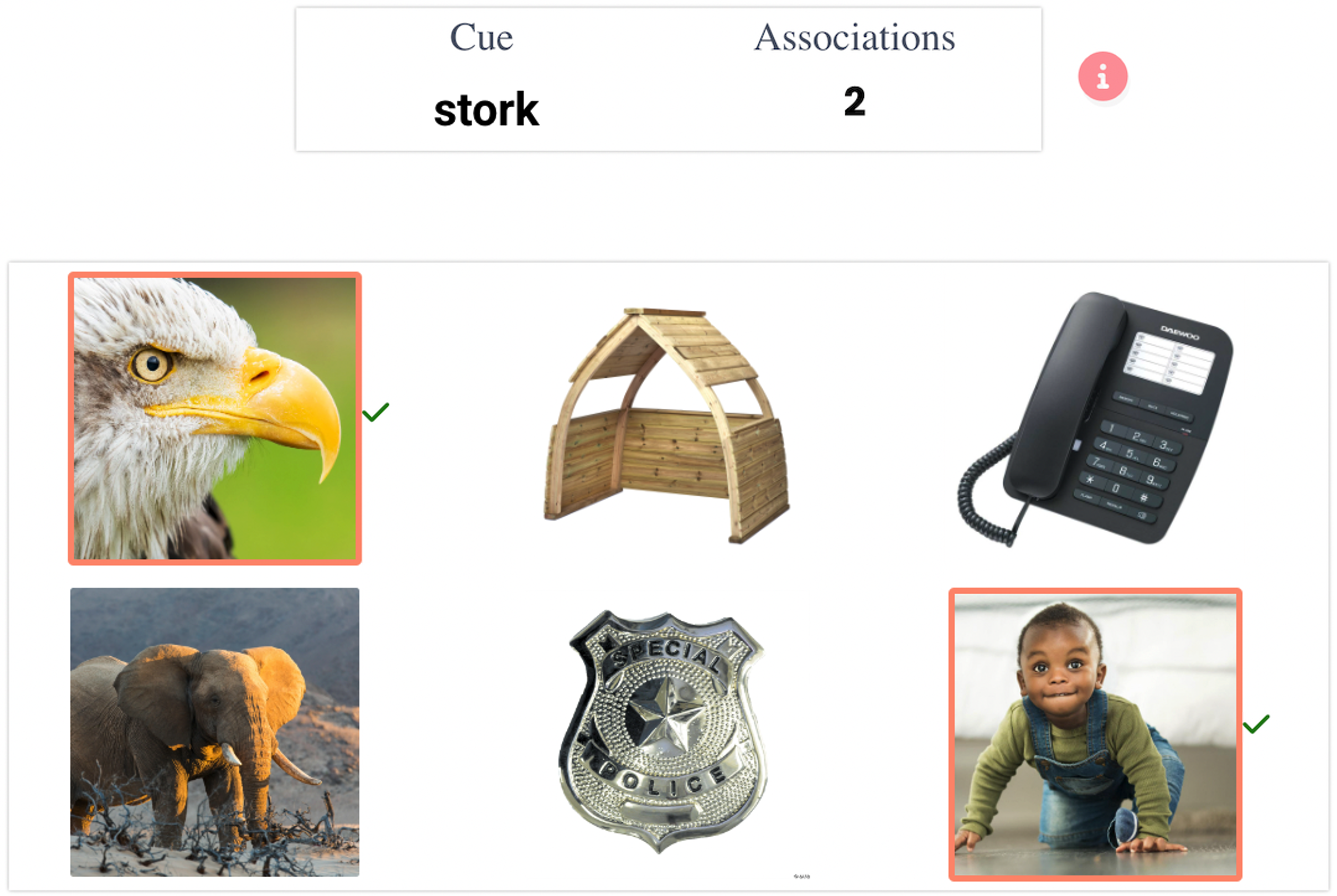}
  \caption{An association from \swowsplit{}, which is automatically extracted based on the SWOW dataset. Annotators receive a cue (e.g., \emph{stork}) and provide three associations. We take the textual annotations, add distractors and extract images for each given association. This association could be solved without the visual information (\emph{stork} is correlated with the concepts of \emph{bird} and \emph{baby}). Cues are assigned in a textual and non-adversarial manner.}
\end{subfigure}
\caption{\gamesplit{} vs. \swowsplit{} generation process.}
\label{fig:swow_vs_winogavil}
\end{figure}

%% file: sections/05B_gvlab_humans_evaluation.tex
\paragraph{Humans.} \phantomsection
\label{sec:human_eval} We sample 10\% of the test sets and validate it with new annotators who were not involved in any previous annotation tasks. We require three different annotators to solve each instance and report their average Jaccard score as the final human prediction. Annotator agreement is measured two different ways: by comparing the Jaccard index of the annotators' selections with the ground-truth labels, and by comparing the Jaccard index between the three annotators' selections. The standard deviations are 6.3, 7.5, and 5, and the Jaccard index is 80, 81, and 89 for the cases with 10-12 candidates, 5-6 candidates, and SWOW, respectively, indicating high agreement.

%% file: sections/05C_gvlab_model_evaluation.tex
\paragraph{Zero-shot models.} We evaluate several diverse state-of-the-art vision-and-language models. 
In all cases described below (except CLIP-ViL), the model encodes the text and the image and produces a matching score for each (cue, image) pair. We take the $k$ (number of associations) images with the top scores (For example, the top $k$=3 model predictions in Figure~\ref{fig:fig2}).\footnote{We ran the zero-shot experiments on a MacBook Pro laptop (CPU) in <6 hours.}

\begin{enumerate}
    \item CLIP \cite{radford2021learning} is pre-trained with a contrastive objective that can be used without directly optimizing for the task. We use four versions of models with different amounts of parameters: RN50, ViT-B/32, ViT-L/14 and RN50x64/14 with 100M, 150M, 430M and 620M parameters respectively (RN50 was used during data collection).
    \item CLIP-ViL \cite{shen2021much}, with 290M parameters, is a pre-trained vision-and-language model that uses CLIP as a visual backbone, rather than CNN based visual encoders that are trained on a small set of manually annotated data. We use the image-text matching objective, where a classification head predicts a score indicating whether the candidate image and the cue match each other. 
    \item ViLT \cite{kim2021vilt}, with 111M parameters, incorporates text embeddings into a Vision Transformer (ViT).
    % \item LiT \cite{zhai2021lit} is used to align vision-text embeddings using contrastive image-text training, used in zero-shot.
    \item X-VLM \cite{zeng2021multi}, with 216M parameters, is pre-trained with multi-grained vision language alignments and fine-tuned for image-text retrieval (Flickr30 \cite{plummer2015flickr30k}) tasks, achieving state-of-the-art results on several benchmarks.
\end{enumerate} 

\paragraph{Supervised models.} We join a line of benchmarks that introduce a test set, without predefined train splits \cite{thrush2022winoground,rudinger2018gender,emelin-sennrich-2021-wino}. We believe that in order to solve associations, a machine must map knowledge to new, unknown cases without extensive training  \cite{mitchell2021abstraction}.  Nonetheless, for completeness, we also consider fine-tuning models on the associations data. We add a binary classifier on top of the pre-trained embeddings to classify whether a given (cue, image) pair is associative or not. We use CLIP (VIT-B/32) model, concatenate the textual cue embedding to the visual image embedding, followed by a classifier that produces a score in $[0,1]$, where 1 is labeled `associative'. We use the Adam   optimizer~\cite{kingma2014adam} with a learning rate of 0.001, batch size of 128, and train for 7 epochs. Since we do not propose a training/validation/test split, we repeat five experiments with different random seeds where we sample a unified training set of 9,326 (cue,image) pairs for both the candidates cases. We then sample a separate test (10\%) and validation (10\%) sets with non-overlapping images, and report the average results, comparing the supervised and zero-shot models on the same sampled test sets.\footnote{Code for reproducing these experiments is available in this \href{https://github.com/WinoGAViL/WinoGAViL-experiments}{link}. We ran the supervised experiments with a single NVIDIA RTX2080 GPU, all experiments ran in <24 hours.}

%% file: figs_and_tables/table_swow_vs_game.tex
\begin{wraptable}{R}{0.5\textwidth}
% \begin{table}
\centering
\caption{Zero-shot models performance on the \swowsplit{} and the \gamesplit{}. Numbers indicates Jaccard score~(0--100\%). Bold numbers indicate best models performances and lowest human performance. The associations collected via the game are difficult for all models to solve.}
\label{tab:table_swow_vs_game}
\begin{tabular}{@{}lccc@{}}
\toprule
Model           & \multicolumn{2}{c}{Game} & SWOW \\ \midrule
\# Candidates   & 10 \& 12        & 5 \& 6       & 5 \& 6  \\ \midrule
CLIP-RN50x64/14 & 38           & 50        & 70   \\
CLIP-VIT-L/14   & 40           & 53        & \textbf{74}   \\
CLIP-VIT-B/32   & 41           & 53        & 74   \\
CLIP-RN50       & 35           & 50        & 73   \\ \midrule
CLIP-ViL        & 15           & 47        & 66   \\
ViLT            & \textbf{52}           & \textbf{55}        & 59   \\
% LiT             & 15           & 31        & 40   \\
X-VLM           & 46           & 53        & 68   \\ \midrule
Humans          & \textbf{90}           & 92       & 95   \\ \bottomrule
\end{tabular}
\end{wraptable}
% \end{table}

%% file: sections/05D_gvlab_results.tex
Zero-shot results on \gamesplit{} and the \swowsplit{} are presented in Table~\ref{tab:table_swow_vs_game}. Table~\ref{tab:human_performance_on_candidates} (Appendix~\ref{sec:appendix}) shows full statistics and performance for the different number of candidates and created associations.

\input{figs_and_tables/table_trainable}

\paragraph{The game allows collection of associations that are easy for humans and challenging for models.} Performance on the data collected via the game is 15--52\% with 10-12 candidates, and 47--55\% with 5-6 candidates. All models' performances are far below human performance (90\% and 92\%, see last row). We highlight that although our rival AI model is CLIP with RN50, the created data is still challenging even for models order-of-magnitude larger. We also see a significant performance drop with most models when increasing the number of candidates without hurting human accuracy, indicating that humans are robust to the increased difficulty level while models struggle with it.

\paragraph{The game creates more challenging associations compared to the SWOW based method.} The highest model performance on the \swowsplit{} is 74\%, and on the \gamesplit{} is 55\%, both with the same number of candidates (5 \& 6). % ADDED - REVISION
CLIP-ViL achieves lower results, especially in the 10 \& 12 case. The reason could be that CLIP-ViL uses the ITM pre-training objective (image-text matching), whereas X-VLM and ViLT are fine-tuned for image-text retrieval. CLIP is also pre-trained, but with a different contrastive pre-training objective that may be more useful for this task. The results indicate the value of our game in collecting associations that are much more challenging than the SWOW-based method.

\paragraph{Training is effective given more distractors.}
Fine-tuning results are presented in Table~\ref{tab:table_trainable}. The relatively low performance indicates that models struggle to capture the information required to solve challenging associations from supervised data. Interestingly, we see that training did not change with 5 \& 6 candidates, but did improve performance by 7\% with 10 \& 12 candidates, indicating that the model is only able to exploit supervised data in particularly hard cases, with lower random chance of success.

%% file: figs_and_tables/table_trainable.tex
% \begin{table}
\begin{wraptable}{R}{0.35\textwidth}
       \caption{Supervised models performance. Results are mean and standard deviation of the Jaccard index of five experiments, each time sampling different test set. Training is effective given more distractors.}
 \label{tab:table_trainable} \begin{tabular}{@{}lcc@{}}
    \toprule
    \#   Candidates & 10 \& 12 & 5 \& 6   \\ \midrule
    Zero-Shot       & 42 $\pm$ 3 & 53 $\pm$ 2 \\
    Supervised      & 49 $\pm$ 3 & 52 $\pm$ 1 \\ \bottomrule
    \end{tabular}
\end{wraptable}
% \end{table}

%% file: sections/05E_model_strength_and_weaknesses.tex
\input{figs_and_tables/strength_and_weaknesses_figure}

\input{figs_and_tables/strength_and_weaknesses_table}
\paragraph{Model performance varies between different association types.} We provide a fine-grained model analysis of different association types. We hypothesize that models perform better on association instances that require direct visual detection, as these models' training objectives are similar to these kind of tasks. We sampled $\sim$1K cue-image pairs of the instances created via the game with 10-12 candidates for analysis. Three of the authors identified the following six categories: (a) \emph{Visually salient}: the cue is the main visually salient item in the image; (b) \emph{Visually non-salient}: the cue is visual but non-salient, and specific to the particular given image; (c) \emph{Concept related}: the cue is related to the image \emph{concept}, not necessarily to the particular given image; (d) \emph{Activity}: the cue is an activity depicted in the image, e.g., the cue is \emph{jumping}, and the image shows people jumping; (e)  \emph{Counting}: the cue is a number or amount of items depicted in the image (e.g., \emph{two} with an image of two people); (f) \emph{Colors}:  the cue indicates a color that is depicted in the image (e.g., \emph{white}); Examples are presented in Figure~\ref{fig:strength_and_weaknesses}. Additional details, annotations guidelines, full examples for each category and screenshots from the annotation task are provided in Appendix~\ref{sec:appendix}, Section~\ref{sec:visually_salient-analysis}. We define the final category as the annotators' majority vote, that was reached in 98\% of the cases, and discarded the other 2\%. 
We evaluate CLIP ViT-B/32 on the association instances and report the accuracy per category which is the proportion of the model success in each given category. Results are presented in Figure~\ref{fig:strength_and_weaknesses}. We find that model performance is highest in the visually salient and colors category, degrades in concept related, and activities, and is much worse in the visually non-salient and counting categories. The results suggest a lack of common sense reasoning capabilities. We release the annotated data in the project website for future research.

%% file: figs_and_tables/strength_and_weaknesses_figure.tex
\begin{figure}
    \centering
    \includegraphics[width=0.8\textwidth]{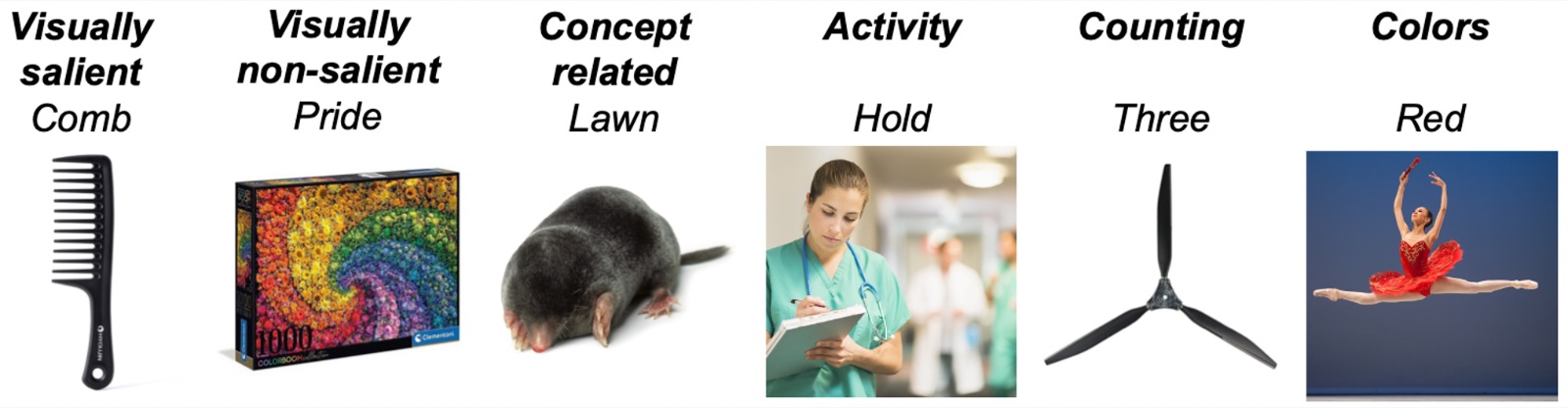}
    \caption{Examples for different association categories and results for each category}
    \label{fig:strength_and_weaknesses}
\end{figure}

%% file: figs_and_tables/strength_and_weaknesses_table.tex
\begin{wraptable}{R}{0.58\textwidth}
\centering
\caption{Results for different association categories and results for each category. The model (CLIP ViT-B/32) is stronger when the cue is visually salient in the image (a), but weaker in the other cases, especially in visually non-salient associations.}
\begin{tabular}{@{}lccc@{}}
\toprule
                  & \# Items & \% Model & \% Humans \\ \midrule
Visually salient     & 67      & 75                                                          & 98                                                              \\
Visually non-salient & 379      & 36                                                            & 93                                                              \\ 
Concept related & 426      & 65                                                            & 92                                                              \\ 
Activity & 24      & 42                                                            & 96                                                              \\ 
Counting & 25      & 36                                                            & 97                                                              \\ 
Colors & 14      & 79                                                            & 96                                                              \\ 

\bottomrule
\newline
\end{tabular}
\label{tab:strength_and_weaknesses}
\end{wraptable}

%% file: sections/05F_image2text.tex
\paragraph{Performance of textual models is close to vision-and-language models, but still far from human.} Another approach for tackling \ouracronym{} is using textual models, when transferring the visual modality to textual modality with image captions, receiving a full-textual dataset. We take OFA \cite{wang2022unifying}, a state-of-the-art image captioning model, and extract image captions for each of the image candidates. We use the three leading models for semantic search in Sentence Transformers \cite{reimers-2019-sentence-bert}, which are Distilled RoBERTa, \cite{Sanh2019DistilBERTAD} and MPNet \cite{song2020mpnet} (two versions, the original model, and a model fine-tuned for semantic search).\footnote{\url{https://www.sbert.net/docs/pretrained_models.html}} Results are presented in Table~\ref{tab:image_captions}. We see that the results are better than chance level, a bit lower than the textual cue and visual candidates' version (ViLT, one line prior to the last), but still far from human performance. These results hint that \ouracronym{} cannot be trivially solved by mapping the images to text.

\input{figs_and_tables/table_image_captions}

%% file: figs_and_tables/table_image_captions.tex
\begin{wraptable}{R}{0.55\textwidth}
\centering
\caption{Results of textual models when using textual image captions for the candidates. ViLT performance (textual cue and visual candidates) performance appear one line prior the the last. Image-to-text might be beneficial, but still far from human performance.}
\label{tab:image_captions} 
\begin{tabular}{@{}lcccccc@{}} \toprule
 Model &  
                                      &                                   & \multicolumn{2}{c}{Game}      & SWOW        \\ \midrule
                                      \# Candidates &  
                                      &                                   & 10 \& 12            & 5 \& 6      & 5 \& 6                \\
                                      \midrule

                                     MPNet          &                       &                                   & 39 &     52        &   72          \\
                                                                          MPNet QA          &                       &                                   &     47 & 55        &   75          \\

                                      Distil RoBERTa          &                       &                                   &     37 & 50        &   65          
\\ \midrule
 ViLT (V\&L)                  &                       &                                  & 52 & 55          & 59\\ 
 Humans                  &                       &                                  & 90 & 92          & 95
 \\ \bottomrule         
\end{tabular}
\end{wraptable}

%% file: sections/03_related_work.tex
\paragraph{Associations and Codenames.} Several works have studied the popular Codenames game in the context of natural language processing \cite{shen2018comparing,kim2019cooperation}, which is also related to works on semantic relatedness \cite{gabrilovich2007computing,strube2006wikirelate,budanitsky2006evaluating,hassan2011semantic}. In the context of associations, a recent work have proposed to use the SWOW resource to evaluate pre-trained word embedding \cite{thawani2019swow}, and some works evaluate models with a CNN-based visual components \cite{de2018visual,de2021visual}. We expand these ideas to evaluate state-of-the-art vision-and-language pre-trained models.

\paragraph{Commonsense.} Commonsense reasoning is a topic with increasing interest lately \cite{choi2022curious}. Many commonsense reasoning tasks have been proposed, both in NLP \cite{saha2021explagraphs,zellers2018swag,zellers2019hellaswag,sap2019atomic,bisk2020piqa,forbes2019neural}, and Computer Vision \cite{fang2020video2commonsense,vedantam2015learning}, including works that require understanding social cues \cite{lei2020more,zellers2019recognition}. In the text domain, a number of Winograd Schema Challenge Datasets have been proposed as alternatives for the Turing test \cite{levesque2012winograd,sakaguchi2020winogrande,kocijan2020review,rudinger2018gender,emelin-sennrich-2021-wino}. In the vision-and-language domain \citet{thrush2022winoground} have proposed a dataset that tests compositional reasoning in vision-and-language models with the task of matching a caption with its correct image. \ouracronym{} also measures vision-and-language reasoning, but focuses on commonsense-based image-cue associations, and primarily serves as a dynamic benchmark as playing the game allows future data collection.

\paragraph{Human-and-model-in-the-loop.} Models are often used in dataset collection to reduce dataset biases or to create adversarial instances \cite{zellers2018swag,zellers2019hellaswag,le2020adversarial,kaushik2019learning,nie2019adversarial}, which might limit the created instances to be effected by the used model. For example, in works that create adversarial visual question answering instances \cite{li2021adversarial,sheng2021human}, human annotators are prompted to fool the model iteratively for each instance, receiving online feedback from the model, and their annotation is allowed to be submitted only after they succeed or after a certain number of trails. In contrast, in our work, the annotators have only one chance to trick the AI model for a given instance. They cannot iteratively `squeeze' the model to produce an adversarial example. Thus, the generated data is less dependent on the particular AI model since the model is only used to motivate the human player to fool it. In particular, we do not use the models' predictions to choose the test set instances. 

\paragraph{Gamification.} Gamification was previously used for several purposes, such as data collection \cite{ipeirotis2014quizz,von2004labeling,eisenschlos2021fool}, education \cite{hays2017game,bustamante2021cultural}, and beat-the-AI tasks for AI model evaluation \cite{bartolo2020beat,attenberg2015beat,chattopadhyay2017evaluating}.   \citet{talmor2022commonsenseqa} proposed a gamification framework to collect question answering instances.
\citet{kiela2021dynabench} proposed a dynamic benchmark that supports human-and-model-in-the-loop.
We propose a game that serves as a dynamic benchmark of vision-and-language associations, gamifying both human interactions with an AI model and human interactions with other humans.

%% file: sections/06_limitations_and_conclusions.tex
\label{sec:limitations_and_conclusions}
Despite our efforts to filter inappropriate concepts and images, some players may feel harmed when they are exposed to new generated cues, or when seeing an image that have passed the automatic and manual filtering. Players are able to mark such cases (with a designated `report' button), leading to immediate removal until further examination. Additionally, players will agree to a consent form when they register. When designing the game, we had several choices to make, including the bonus reward and the AI model interaction. Future work will thoroughly explore the impact of these choices.

We introduced an online game to collect challenging associations. We demonstrated its effectiveness by collecting a dataset that it is easy for humans and challenging for state-of-the-art models. We provided an extensive evaluation of the game and collected dataset. We hope the \ouracronym{} benchmark will drive the development of models with better commonsense and association abilities.

%% file: sections/acknowledgement.tex
We would like to thank Moran Mizrahi for a feedback regarding the players survey. We would also like to thank Jaemin Cho, Tom Hope, Yonatan Belinkov, Inbal Magar and Aviv Shamsian. This work was supported in part by the Center for Interdisciplinary Data Science Research at the Hebrew University of Jerusalem, and a research grant no. 2088 from the Israeli Ministry of Science and Technology.

%% file: sections/paper_checklist.tex
% %%% BEGIN INSTRUCTIONS %%%
% The checklist follows the references.  Please
% read the checklist guidelines carefully for information on how to answer these
% questions.  For each question, change the default \answerTODO{} to \answerYes{},
% \answerNo{}, or \answerNA{}.  You are strongly encouraged to include a {\bf
% justification to your answer}, either by referencing the appropriate section of
% your paper or providing a brief inline description.  For example:
% \begin{itemize}
%   \item Did you include the license to the code and datasets? \answerYes{See Section~\ref{gen_inst}.}
%   \item Did you include the license to the code and datasets? \answerNo{The code and the data are proprietary.}
%   \item Did you include the license to the code and datasets? \answerNA{}
% \end{itemize}
% Please do not modify the questions and only use the provided macros for your
% answers.  Note that the Checklist section does not count towards the page
% limit.  In your paper, please delete this instructions block and only keep the
% Checklist section heading above along with the questions/answers below.
% %%% END INSTRUCTIONS %%%

\begin{enumerate}

\item For all authors...
\begin{enumerate}
  \item Do the main claims made in the abstract and introduction accurately reflect the paper's contributions and scope?
    \answerYes{}
  \item Did you describe the limitations of your work?
    \answerYes{See Section~\ref{sec:limitations_and_conclusions}}
  \item Did you discuss any potential negative societal impacts of your work?
    \answerYes{See Section~\ref{sec:limitations_and_conclusions}}
  \item Have you read the ethics review guidelines and ensured that your paper conforms to them?
    \answerYes{}
\end{enumerate}

\item If you are including theoretical results...
\begin{enumerate}
  \item Did you state the full set of assumptions of all theoretical results?
    \answerNA{}
	\item Did you include complete proofs of all theoretical results?
    \answerNA{}
\end{enumerate}

\item If you ran experiments (e.g. for benchmarks)...
\begin{enumerate}
  \item Did you include the code, data, and instructions needed to reproduce the main experimental results (either in the supplemental material or as a URL)?
    \answerYes{\ourwebsite{}}
  \item Did you specify all the training details (e.g., data splits, hyperparameters, how they were chosen)?
    \answerYes{}
	\item Did you report error bars (e.g., with respect to the random seed after running experiments multiple times)?
    \answerYes{Reporting results average and standard-deviatio Table~\ref{tab:table_trainable}, Table~\ref{sec:human_eval}}
	\item Did you include the total amount of compute and the type of resources used (e.g., type of GPUs, internal cluster, or cloud provider)?
    \answerYes{}
\end{enumerate}

\item If you are using existing assets (e.g., code, data, models) or curating/releasing new assets...
\begin{enumerate}
  \item If your work uses existing assets, did you cite the creators?
    \answerYes{Section~\ref{sec:swow}}
  \item Did you mention the license of the assets?
    \answerYes{Section~\ref{sec:swow}}
  \item Did you include any new assets either in the supplemental material or as a URL?
    \answerYes{\ourwebsite{}}
  \item Did you discuss whether and how consent was obtained from people whose data you're using/curating?
    \answerYes{}
  \item Did you discuss whether the data you are using/curating contains personally identifiable information or offensive content?
    \answerYes{}
\end{enumerate}

\item If you used crowdsourcing or conducted research with human subjects...
\begin{enumerate}
  \item Did you include the full text of instructions given to participants and screenshots, if applicable?
    \answerYes{Section~\ref{sec:quals}}
  \item Did you describe any potential participant risks, with links to Institutional Review Board (IRB) approvals, if applicable?
    \answerNA{}
  \item Did you include the estimated hourly wage paid to participants and the total amount spent on participant compensation?
    \answerYes{Section~\ref{sec:human_evaluation_and_annotation}}
\end{enumerate}

\end{enumerate}

%% file: sections/99_appendix.tex
\label{sec:appendix}

\subsection{Dataset Supplementary Materials}
\label{sec:dataset_supp_materials}
\input{sections/supp_dataset_appendix}

\subsection{Reasoning Skills}
Table~\ref{tab:reasoning_skills} lists the full reasoning and observed patterns annotated to solve the \gamesplit{} (\S\ref{sec:reasoning_skills}), and Figure~\ref{fig:examples_visual_reasoning} shows an example of each visual pattern we annotated. 
\input{figs_and_tables/table_reasoning_skills}
\input{figs_and_tables/fig_visual_associations_examples}

\subsection{Human Annotation}
\input{sections/99_appendix_human_annotation}

\subsection{Full Results}
\input{figs_and_tables/table_human_performance_on_candidates}

\subsection{Model Analysis on Different Association Types}
\label{sec:visually_salient-analysis}
A sample of 1,000 associations were annotated by three different annotators. We defined the final category as the annotators' majority vote, that was reached in 98\% of the cases, and discarded the other 2\%. We reported the accuracy per category, which is the proportion of the model success in each given category. 
The full annotation guidelines are available in the following link: \url{https://github.com/WinoGAViL/WinoGAViL-experiments/tree/main/assets/association_types_annotations_guidelines.pdf}
The annotated data is available in the following link: \url{https://github.com/WinoGAViL/WinoGAViL-experiments/blob/main/assets/model_analysis_on_different_association_types.csv}
We show examples of the annotated data categories in Figures~\ref{fig:visually_salient},\ref{fig:visually_non_salient},\ref{fig:grid_concept},\ref{fig:grid_activity}, \ref{fig:grid_counting}, \ref{fig:grid_colors}. A screenshot from the annotation task is presented in Figure~\ref{fig:mturk_association_type_task}.

\input{figs_and_tables/model_analysis_grid}
\clearpage
\newpage
\clearpage
\newpage
\input{figs_and_tables/fig_example_mturk_association_type}

\subsection{Multimodal Evaluation}
\input{sections/99A_swow_multimodal}

%% file: sections/supp_dataset_appendix.tex
\begin{enumerate}
  \item Dataset documentation, metadata, and download instructions: \url{https://winogavil.github.io/download}.
  \item Intended uses: we hope our benchmark will be used by researchers to evaluate machine learning models. We hope that our benchmark will be played by users, leading to new associations collection.
  \item Author statement: We bear all responsibility in case of violation of right in using our benchmark. 
  \item Licenses: Code is licensed under the MIT license \url{https://opensource.org/licenses/MIT}. Dataset is licensed under CC-BY 4.0 license \url{https://creativecommons.org/licenses/by/4.0/legalcode}.
  \item Hosting \& preservation: our website is deployed and all data is accessible and available. We encourage researchers to send us model predictions on the created test sets. We will update a model and players leader-board with this results periodically. 
  \item Code repository: \url{https://github.com/WinoGAViL/WinoGAViL-experiments}
\end{enumerate}

%% file: figs_and_tables/table_reasoning_skills.tex
\begin{table}[!htb]
\caption{Some of the observed patterns and reasoning skills required to solve \ouracronym{} associations. Each association instance may require multiple skills}
\label{tab:reasoning_skills}
\resizebox{\textwidth}{!}{
\begin{tabular}{@{}lllll@{}} \toprule
\textbf{Skill} & \textbf{Observed Pattern}                                & \textbf{Description}                                                                                                    & \textbf{Example} & \textbf{\%}                                                                                                                                              \\ \midrule
\multirow{20}{*}{Non-Visual} & {Kind-Of}             & \textcolor{red}{Cue} is a kind of \textcolor{mygreen}{Association}                                                                                   & a \textcolor{red}{bathtub} is a \textcolor{mygreen}{shower}          & {4\%}                                                                                                                       \\
                                     & & \textcolor{mygreen}{Association} and \textcolor{red}{Cue} are kinds of \textcolor{blue}{Something}                                                                     & a \textcolor{red}{croissant} \& \textcolor{mygreen}{bread} are \textcolor{blue}{pastries} &                                                                                                                     \\ 
                                     \cmidrule(l){2-5}
 &{Attribute}           & \textcolor{red}{Cue} has attributes of \textcolor{mygreen}{Association}                                                                              & \textcolor{red}{iguana} has \textcolor{mygreen}{green} color  & {14\%}                                                                                                                               \\
                                     & & \textcolor{red}{Cue} is \textcolor{mygreen}{Association}                                                                                             & \textcolor{red}{miners} are \textcolor{mygreen}{dirty} &                                                                                                                                     \\ \cmidrule(l){2-5}
&{Use-Of}              & \textcolor{red}{Cue} uses the \textcolor{mygreen}{Association}                                                                                       & \textcolor{red}{miner} uses \textcolor{mygreen}{tractor}   & {9\%}                                                                                                                                  \\
                                     & & \textcolor{mygreen}{Association} is used in relation to \textcolor{red}{Cue}                                                                         & \textcolor{red}{tupperware} is used to store \textcolor{mygreen}{food} &                                                                                                                     \\ \cmidrule(l){2-5}
&{General   Knowledge} & \textcolor{red}{Cue} is a name for \textcolor{mygreen}{Association}                                                                & \textcolor{red}{ford} is a name of a \textcolor{mygreen}{car}  & {13\%}                                                                                                                   \\
 & & \textcolor{mygreen}{Association} is used in a relation to \textcolor{red}{Cue}                                                                & \textcolor{mygreen}{oats} for \textcolor{red}{horses} increase their performance  &                                                                                                                  \\
                          \cmidrule(l){2-5}
&{Word Sense Meaning}  & \multirow{4}{*}{\textcolor{red}{Cue} has word sense meaning with \textcolor{mygreen}{Association}}                                                   & \begin{tabular}[c]{@{}l@{}}skin $\leftrightarrow$ object: \textcolor{red}{iguanas} have \textcolor{mygreen}{scale}s, \\ but it is also used to measure weight\end{tabular}   & {3\%}      \\   \cmidrule(l){4-4} 
&   &                                                     & \begin{tabular}[c]{@{}l@{}}visible trail $\leftrightarrow$ body part: \textcolor{red}{comets} have \textcolor{mygreen}{tail}, \\ but it is also an animal body part \end{tabular}   & {3\%}      \\   \cmidrule(l){2-5}

&{Locations}           & The location of a \textcolor{red}{Cue} is \textcolor{mygreen}{Association}                                                                           & \textcolor{red}{comet} is in the \textcolor{mygreen}{sky}  & {5\%}                                                                                                                              \\
& & \textcolor{red}{Cue} and \textcolor{mygreen}{Associations} are located \textcolor{blue}{Somewhere}                                                                           & \textcolor{red}{polar bears} live in an \textcolor{mygreen}{ice} environment &                                                                                                                                \\ \cmidrule(l){2-5}
&{Outcome}             & \textcolor{red}{Cue} is an outcome of \textcolor{mygreen}{Association}                                                                               & \textcolor{red}{oboe} creates \textcolor{mygreen}{music}    & {6\%}                                                                                                                                 \\
& & \textcolor{mygreen}{Association} is an outcome of \textcolor{red}{Cue}                                                                              & \textcolor{mygreen}{birth} \& \textcolor{mygreen}{baby} is the outcome of a \textcolor{red}{pregnancy}    &                                                                                                                                                        \\  \midrule   
                          
\multirow{10}{*}{Visual}&{Activity}  
                                     & \multirow{1}{*}{\textcolor{mygreen}{Association}s perform a \textcolor{red}{Cue}} in the image                                                             & \textcolor{mygreen}{deer} \& \textcolor{mygreen}{snowman} looks like they \textcolor{red}{stare} (Figure \ref{fig:examples_visual_reasoning}b)  &  {6\%}                                                                                                  \\ \cmidrule(l){2-5}%\midrule%\bottomrule  
                                     
&{Humor/Sarcasm}             & \textcolor{red}{Cue} is related to \textcolor{mygreen}{Association} in a funny way & \textcolor{red}{pigpen} is a dirty place, \textcolor{mygreen}{tide} can make it cleaner (Figure \ref{fig:examples_visual_reasoning}e) & {1\%}                                                                                                                                 \\
                                     && & a \textcolor{mygreen}{man} that looks neglected is described as \textcolor{red}{trim}  &                                                                                                    \\ \cmidrule(l){2-5}%\bottomrule  
&{Analogy}             & \begin{tabular}[c]{@{}l@{}}\textcolor{red}{Cue} can be seen/used like/with \textcolor{mygreen}{Association},\\
although they are from a different concept map \end{tabular}                                                                             & \textcolor{mygreen}{TV antenna} looks like a \textcolor{red}{horn} (Figure \ref{fig:examples_visual_reasoning}d)    & {4\%}                                                                                                                                                                                                                     \\\cmidrule(l){2-5}
&{Visual Similarity}             & \textcolor{mygreen}{Association} is visually similar to the \textcolor{red}{Cue}                                                                               & The \textcolor{mygreen}{sponge} shape is similar to a \textcolor{red}{box} (Figure \ref{fig:examples_visual_reasoning}a)    & {20\%}                                                                                                                                                                     \\ \cmidrule(l){2-5}%\bottomrule  
&{Abstraction}    & \textcolor{red}{Cue} is related to \textcolor{mygreen}{Association} in an abstract way & \textcolor{red}{discovery} is when a \textcolor{mygreen}{bulb} turns on (I got it!) (Figure \ref{fig:examples_visual_reasoning}c)  & {5\%}        

                                    \\ \cmidrule(l){2-5}%\bottomrule  
&{Generalization}    & \multirow{2}{*}{ -}                                                                                                                                          & \textcolor{mygreen}{bread dough} becomes \textcolor{red}{fresh} bread when baked     & {8\%}                                                                                                  \\
                                     &&                                                                                                                & \textcolor{mygreen}{raven} is a bird that can be found in a \textcolor{red}{backyard} &                                                                                                       \\ \bottomrule  
\end{tabular}
}
\end{table}

%% file: figs_and_tables/fig_visual_associations_examples.tex
\begin{figure}[!htb]
\centering
\begin{subfigure}[b]{0.425\textwidth}
   \includegraphics[width=1\linewidth]{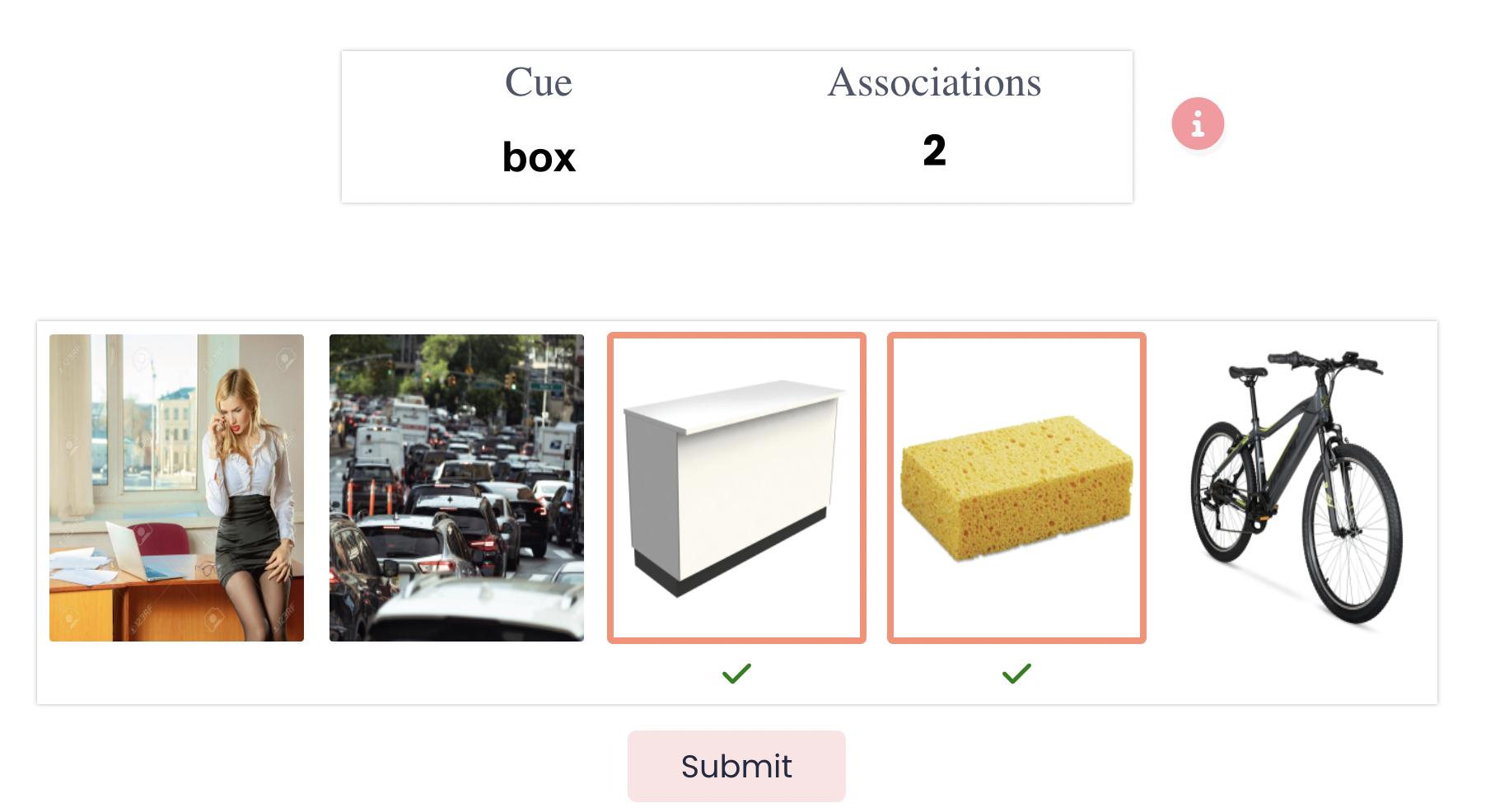}
   \caption{Visual similarity}
\end{subfigure}

\begin{subfigure}[b]{0.425\textwidth}
   \includegraphics[width=1\linewidth]{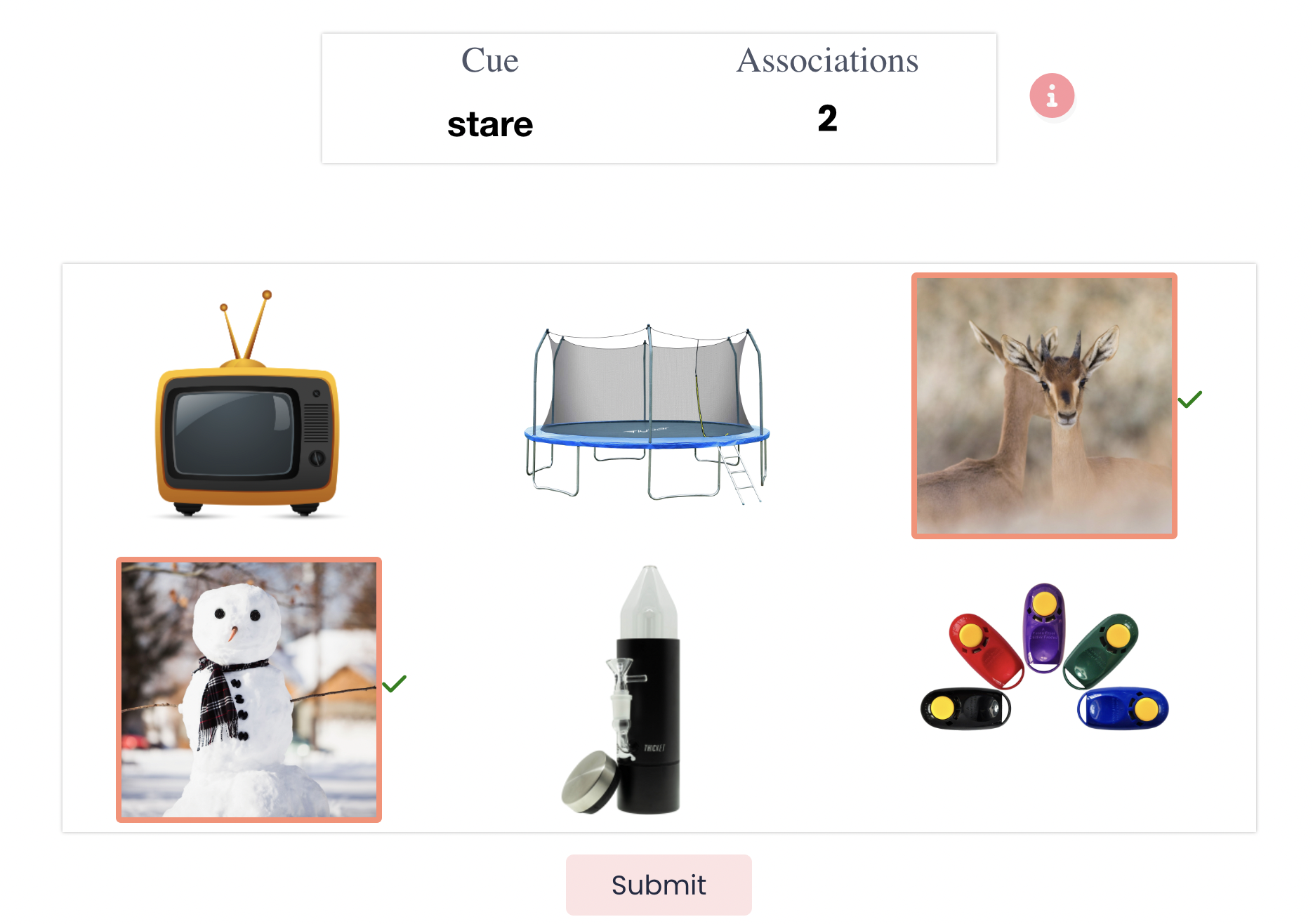}
   \caption{Activity}
\end{subfigure}

\begin{subfigure}[b]{0.425\textwidth}
   \includegraphics[width=1\linewidth]{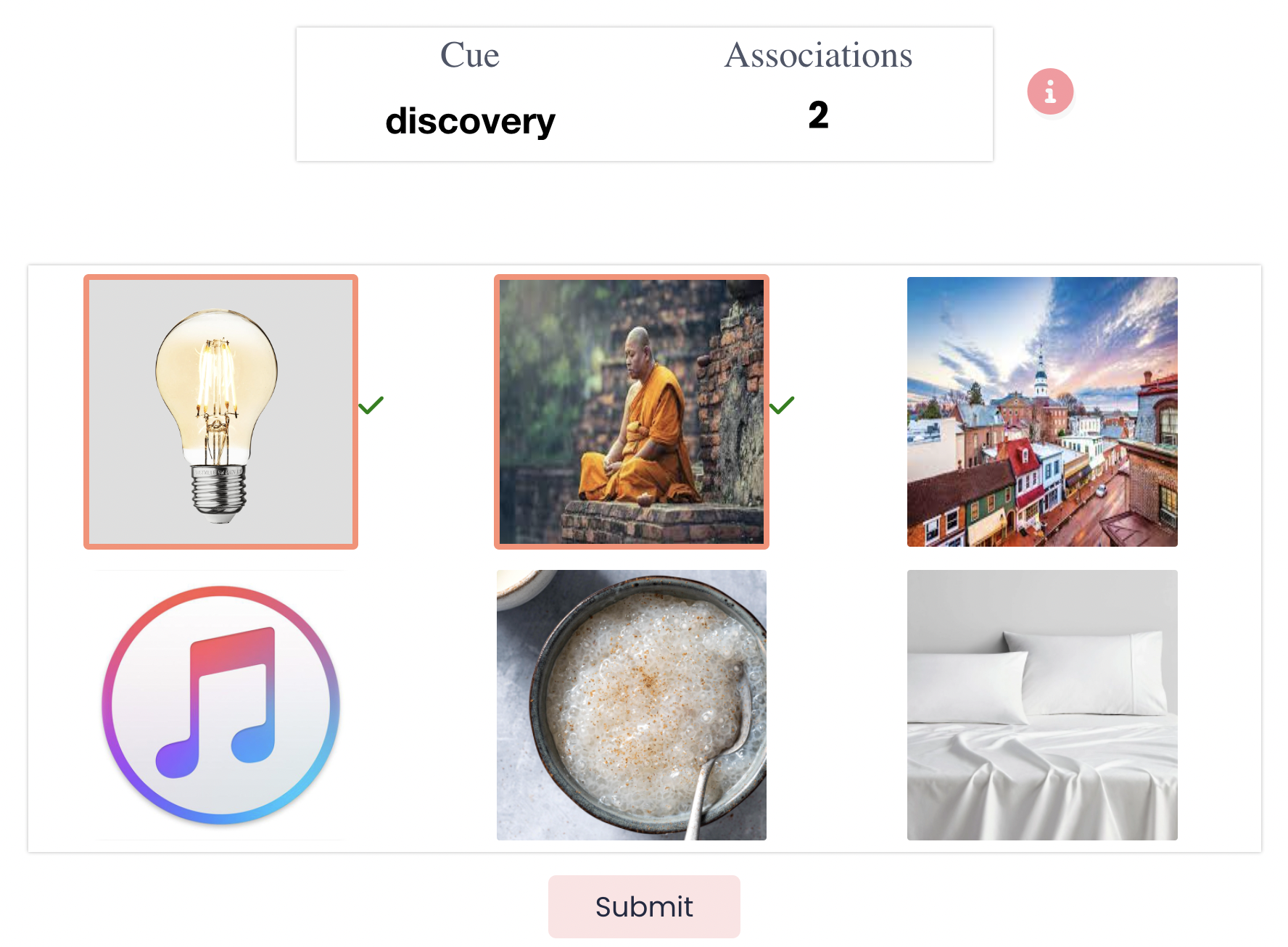}
   \caption{Abstraction}
\end{subfigure}

\begin{subfigure}[b]{0.425\textwidth}
   \includegraphics[width=1\linewidth]{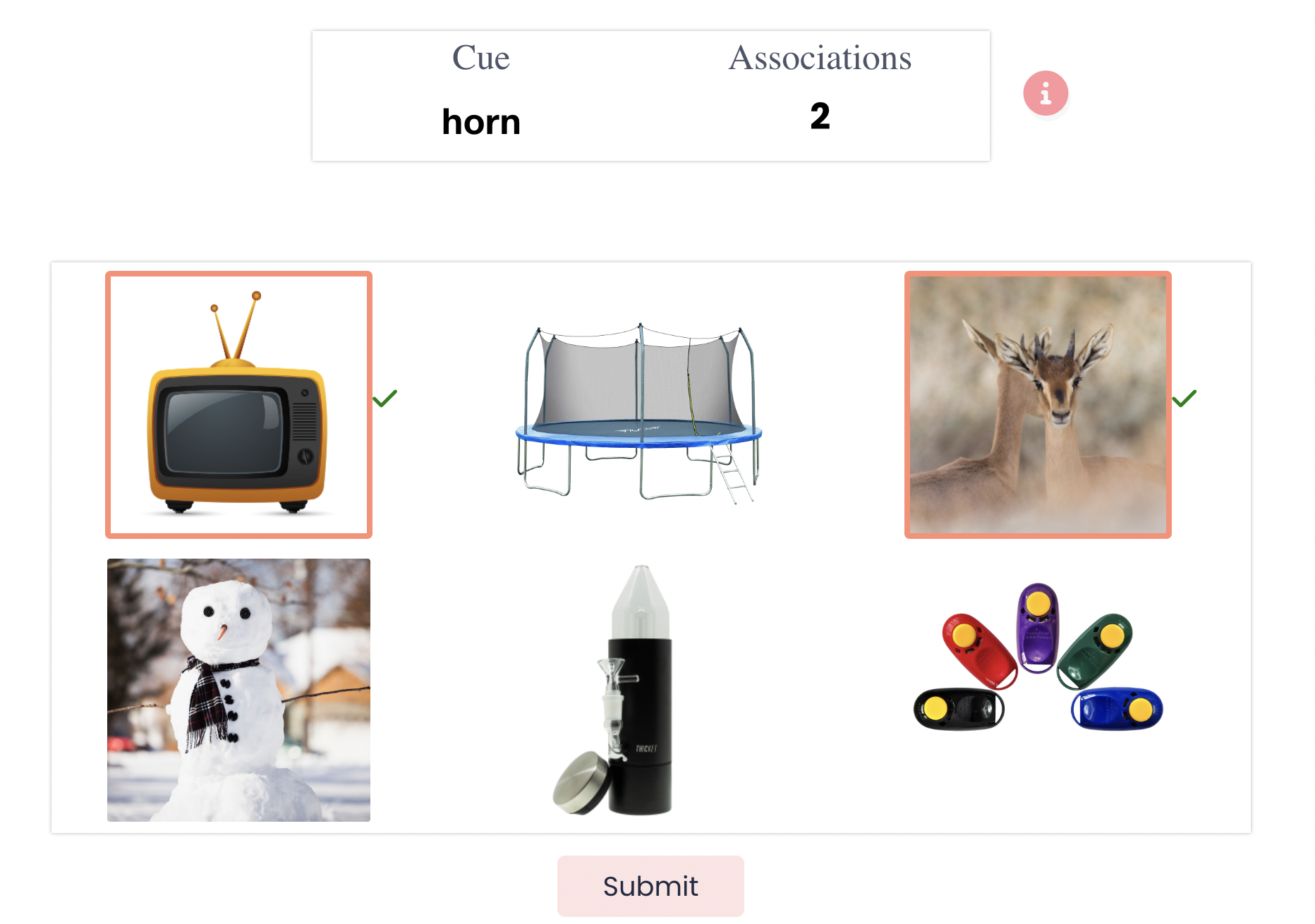}
   \caption{Analogy}
\end{subfigure}

\begin{subfigure}[b]{0.425\textwidth}
   \includegraphics[width=1\linewidth]{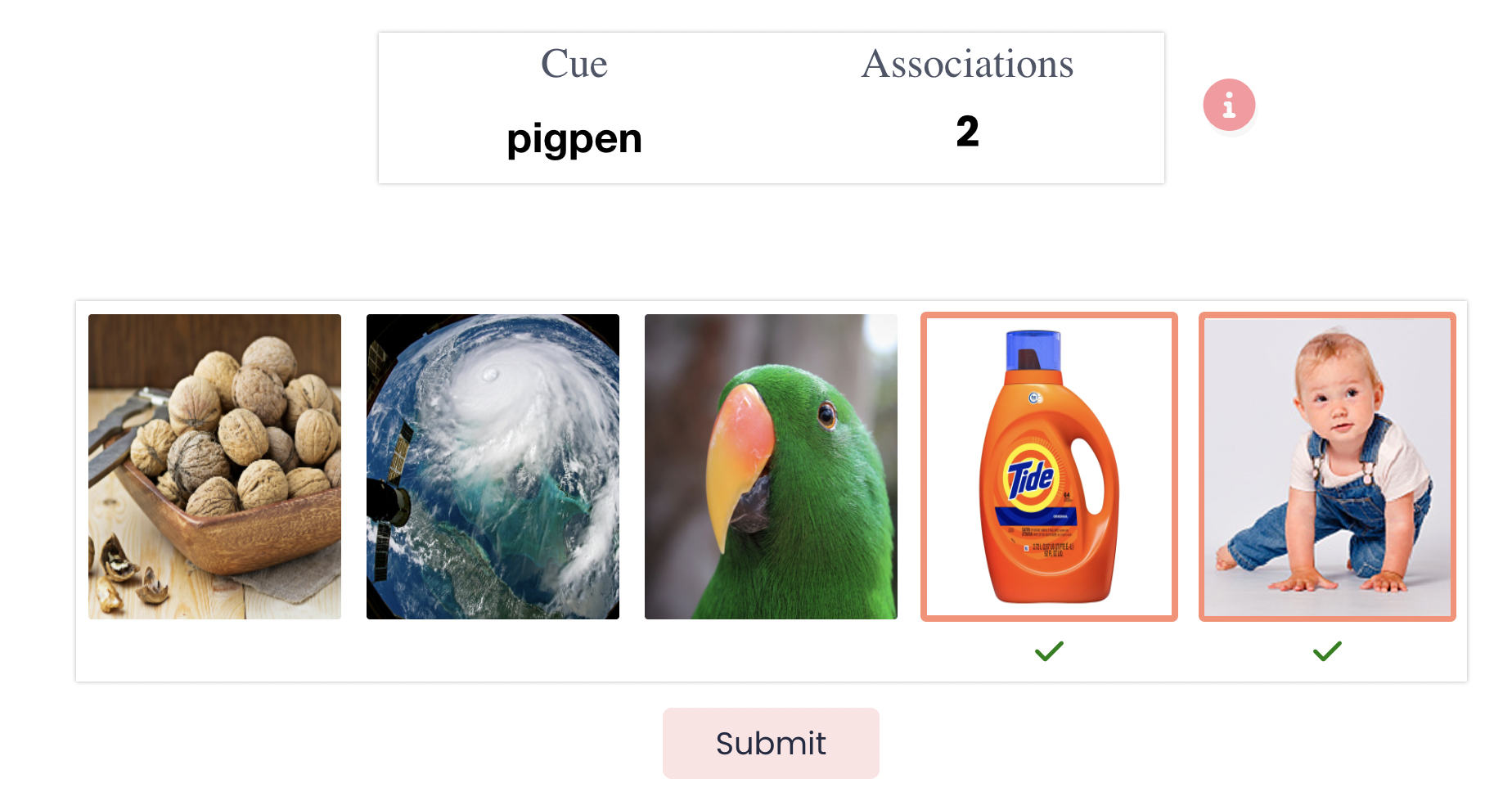}
   \caption{Sarcasm}
\end{subfigure}

\caption{Visual Reasoning Skills Examples}
\label{fig:examples_visual_reasoning}
\end{figure}
% \FloatBarrier

%% file: sections/99_appendix_human_annotation.tex
Figure~\ref{fig:mturk_ui_examples} shows an example of the Mechanical Turk user-interface. Section~\ref{sec:quals} describe the annotator qualifications we required. Section~\ref{sec:bonus} describes the designed bonus reward, aiming to receive generated data that is challenging for models and easy for humans. Section~\ref{sec:players_feedback} describes the player feedback we collected. Finally, Section~\ref{sec:additional_analysis} describes additional analysis such as players statistics and the generated textual cues analysis. 

\input{figs_and_tables/fig_mturk_ui_example}

\subsection{Qualifications} 
\label{sec:quals}
The basic requirements for our annotation task is percentage of approved assignments above 98\%, more than 5,000 approved HITs, the location from the US, UK, Australia or New Zealand. To be a `solver' or a `spymaster', we required additional qualification tests: We selected 10 challenging examples from SWOW based dataset as qualification test. In each qualification test, a new worker entered demographic information: age, gender, level of education and whether he is a native English speaker. To be qualified as a `solver', we accepted annotators that received a mean jaccard score over 80\%. To be qualified as a `creator', we require ``fool-the-AI'' score above 40\%, and ``solvable-by-humans'' score above 80\%. To obtain ``solvable-by-humans'' score, we sent the created associations to solvers (who have passed to solve qualification). The players received instructions, presented in \ref{fig:instructions} and could do an interactive practice in the project website.\footnote{\url{https://winogavil.github.io/beat-the-ai}}. We do not collect or publish players personal information. We presented anonymous demographic statistics, and we do not publish the demographic information.

\input{figs_and_tables/instructions}

\subsubsection{Bonus Reward}
\label{sec:bonus}
If the score is between [50,60), the bonus is 0.03\$. If the score is between [60,67), the bonus is 0.07\$. If the score is between [67,80), the bonus is 0.18\$. Finally, if the score is at least 80, the bonus is 0.27\$. The payment can thus reach up to 0.61\$ for a single annotation when creating two cues for the same image instances that completely fool the AI model and are still solvable by humans.

\subsubsection{Players Feedback}
\label{sec:players_feedback}
\input{sections/99B_players_feedback}

\subsection{Additional Analysis}
\label{sec:additional_analysis}
\paragraph{Annotators statistics.}
Table~\ref{tab:workers_statistics} in Appendix~\ref{sec:appendix} presents statistics for the Amazon Mechanical Turk workers that were involved in \ouracronym{} annotation, both as spymasters and as solvers. A total of 58 crowd workers, mostly English native speakers ($\geq$95\%), of a variety of ages (26--65), genders, and levels of education (high school to graduate school). Figure~\ref{fig:spymasters_stats} in Appendix~\ref{sec:appendix} presents the spymaster's score plots, which include the number of annotations, fool-the-AI score, and solvable-by-humans score for each spymaster.

\input{figs_and_tables/tables_workers_statistics}

\input{figs_and_tables/fig_spymasters_statistics}
\input{figs_and_tables/fig_statistics}

\paragraph{Generated cues statistics.} For the final 3,568 test instances, 2,215 different cues were collected. We measure the concreteness of cue words using the concreteness dataset described \cite{brysbaert2014concreteness}, in which human annotated concreteness scores on a scale of 1-5 were collected. This dataset covers over 88\% of the collected cues, indicating a 12\% upper bound for out-of-vocabulary words. We see a diversity of both abstract and concrete generated cues in Figure~\ref{fig:concreteness}, Appendix~\ref{sec:appendix}. Additionally, we measure how often different annotators compose the same cues for the same group of images. Since we asked three different annotators to provide two different cues for each group of images, we have six annotations for each image group. We find that almost always (98\%) they combine different cues. 

\input{figs_and_tables/fig_concreteness}

%% file: figs_and_tables/fig_mturk_ui_example.tex
\begin{figure}[!htb]
\centering
\begin{subfigure}[b]{0.9\textwidth}
   \includegraphics[width=1\linewidth]{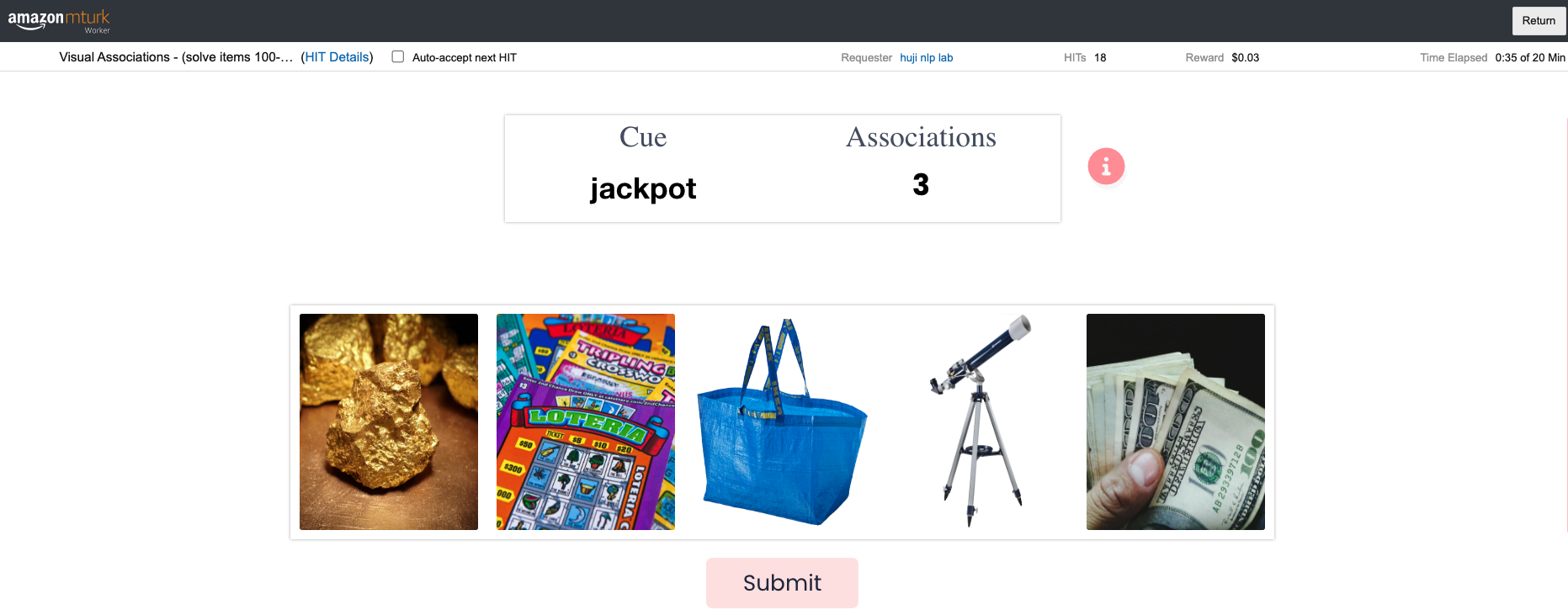}
   \caption{A screenshot from a solver screen in Amazon Mechanical Turk. Basic payment is 0.03\$.}
\end{subfigure}

\begin{subfigure}[b]{0.9\textwidth}
   \includegraphics[width=1\linewidth]{images/example_create.png}
   \caption{A screenshot from a spymaster screen in Amazon Mechanical Turk. Basic payment is 0.07\$.}
\end{subfigure}

\caption{Examples of the Mechanical Turk user-interface, which referred the crowd workers to the \ouracronym{} website}
\label{fig:mturk_ui_examples}
\end{figure}
% \FloatBarrier

%% file: figs_and_tables/instructions.tex
\begin{figure}[!htb]
\centering
\newcommand{\figlen}[0]{\columnwidth}
    \includegraphics[width=0.9\figlen]{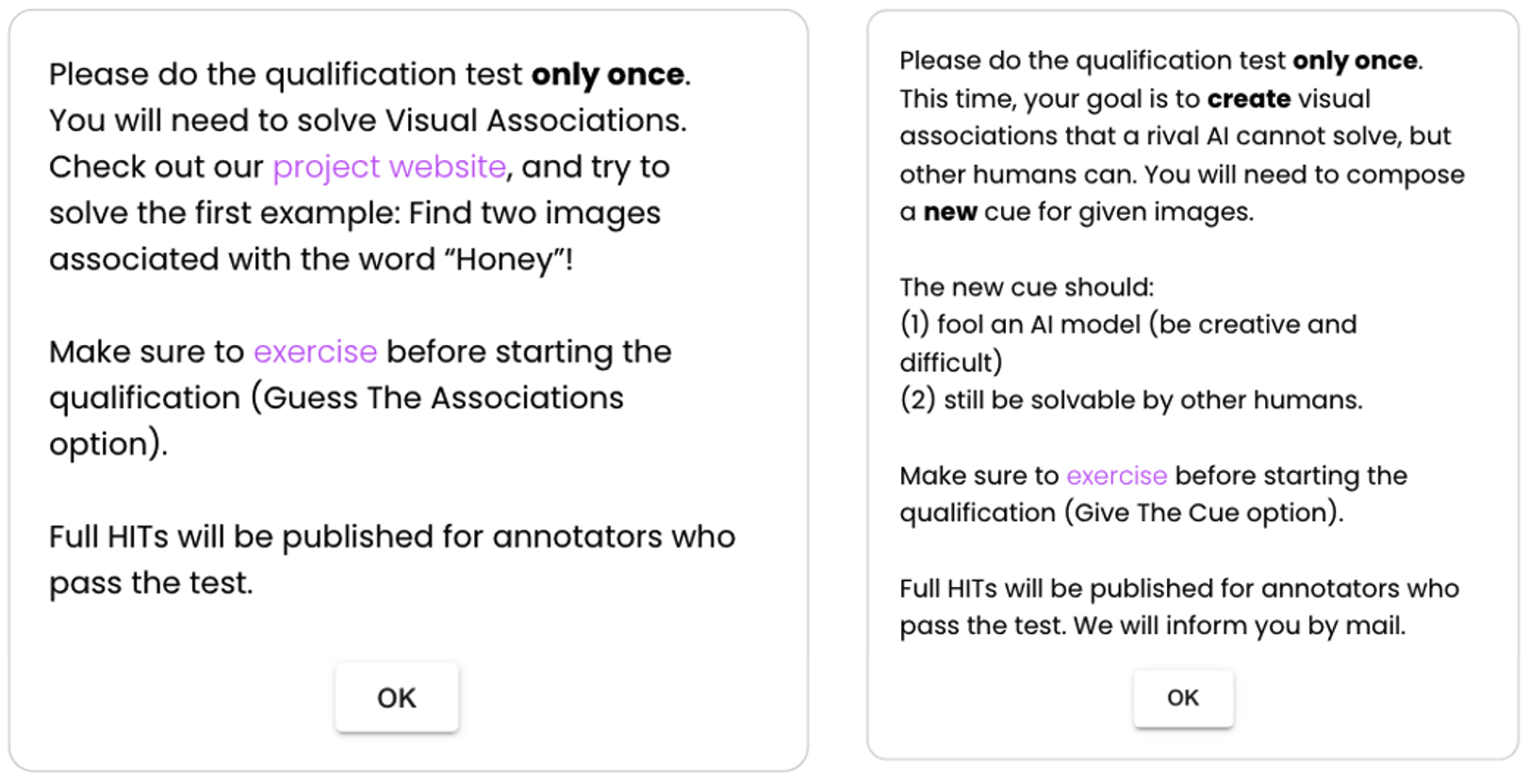}\\
    \caption{A screenshot of the instructions given to the annotators.}
    \label{fig:instructions}
\end{figure}

%% file: sections/99B_players_feedback.tex
Here we list some of the open text feedback we received from our crowd workers. It is not cherry-picked - we chose five representative responses with positive and negative insights.

Q: Describe what did you like and dislike while performing the task. \\
Spymasters:
\begin{enumerate}
    \item I liked the chance to improve my creativity and brainstorm. It was fun.
    \item I liked the mental challenge, especially on the larger 10-12 ones. It was frustrating when the AI clearly guessed and got it right on the 5-6. 
    \item I liked that I got immediate feedback and it was something different than what I usually do on mturk. I did not like that sometimes it seemed like the objects had nothing in common and it took me too long to think of a word to try and associate the objects.
    \item I liked that it was a very creative-focused task, even more so on the creator’s side. It was fun to think of what I could come up with to link these words/images and fool the AI/other people.
    \item Creating was exponentially harder for me than the solving. I felt frustration and I kind of felt stupid because I struggled with it. (But the solving was a blast.)
\end{enumerate}
 
Solvers:
\begin{enumerate}
    \item I liked how easy and straightforward they were, and that they were also super fun and different from other typical HITs I have done. The only thing I disliked was probably the pay but it was not a big deal.
    \item I like the fact that I got to be creativity. Nothing to dislike about this task.
    \item I liked that the correct answers were sometimes abstract and required a little thinking.
    \item I liked that it was a puzzle. I really enjoy puzzles. I did not like that some of them seemed unsolvable. But all in all, I enjoyed it and did much more than I usually do.
    \item I liked trying to figure out what the creator was thinking
\end{enumerate}

Q: Are there additional reasoning skills you feel that were required from you?

Spymasters:
\begin{enumerate}
    \item I find things like common sense and general knowledge mattered less for creating than when solving, because the AI was very good at cracking anything using general knowledge. You had to go more for abstract, metaphorical, or otherwise really ‘out there’ associations to get past it.
\end{enumerate}
 
Solvers:
\begin{enumerate}
    \item This is probably covered under “general knowledge” but I found that a lot of answers required a basic understanding of Pop Culture references.
    \item Luck, of course, but also a fair bit of pop culture wisdom, which is separate from general knowledge.
    \item Seeing a different perspective.
\end{enumerate}

Q: Did seeing the model’s predictions affect you in any way? If so, how? (For spymasters only)

\begin{enumerate}
    \item I was impressed at some of the AI ideas, admired the programmers and learning.
    \item Yes, it helped but it was also kind of discouraging as it seemed like the AI was able to guess nearly all of my associations, which made me feel like I had even more limited options.
    \item I used the model’s guesses to make my associations better. I went after associations that the model frequently got wrong.
    \item Yes, it either increased my confidence or made me think harder about cues.
    \item Sometimes the model was very off especially in detecting emotions.
\end{enumerate}

Q: Have you been affected by the performance bonus? In what way? (For spymasters only)

\begin{enumerate}
    \item It was nice to have a little extra pay. It helped to keep my motivation up when it was hard to come up with connections.
    \item The bonus did make me sometimes give up on making a “good” cue and make a “performance” cue. Performance cue being a cue that utilizes a quirk of the AI that I know and almost guarantee that it will get wrong and will generally be easy for humans to guess. But it’s not a creative or interesting cue. Notable words are human, male and female or sometimes features like eyes, noses, ears, hands, etc.
    \item Yes, it made me try harder to fool the AI.
    \item The performance bonus motivated me to try harder to beat the AI, so I could justify the time investment.
    \item Not really, it wasn’t enough of a bonus for me to be motivated to do more
\end{enumerate}

Q: Anything else that you want to say? 

\begin{enumerate}
    \item I enjoyed this a lot and hope to participate in similar tasks for you in the near future!
    \item It was fun and I hope the best for this project! If you make an online game I would 100\% suggest a leaderboard for “creators” for people to create the cues. Introduce categories so people can focus on specific things. If you’re also so inclined, build something to work with Twitch.tv so streamers can play with their audience. There are some pictionary like games that do this where the streamer draws and the people in chat try to guess.
    \item This would be a super interesting online if you include things like leaderboards for creators, categories, more images (although be sure to get rights to images!) and letting people rate the cues. I can definitely see game like this being popular with streamers on Twitch.tv to play with their audience (streamer https://twitch.tv/itshafu is pretty known to like games like these and sometimes streams her playing code names with other streamers) or with a group of people online.
    \item This was something different to do and was fun, thank you for the opportunity. I also really appreciated how you communicated with us!
    \item I liked creating, more than solving, even though I think I was a better solver than creator; I’m hoping to read the paper that results from this research. 
\end{enumerate}

%% file: figs_and_tables/tables_workers_statistics.tex
\begin{table}[!htb]
\centering
\caption{\ouracronym{} Workers Statistics}
\begin{tabular}{@{}lll@{}}
\toprule
                           & Solvers       & Creators                                                                           \\ \midrule
\# Workers                 & 41            & 18                                                                                 \\
\# Avg. Annotations        & 567           & 332                                                                                \\
\% Avg. Performance (5-6 candidates split)       & 85.1          & \begin{tabular}[c]{@{}l@{}}fool-the-AI:   50\\ solvable-by-humans: 83\end{tabular} \\
Avg. Age         & 41 $\pm$10 & 43 $\pm$9                                                                    \\
\# High School Education           & 13      & 6                                                                           \\
\# Bachelor Education           & 19      & 11                                                                           \\
\# Master Education           & 8      & 1                                                                           \\

\% Native English Speakers & 98            & 95                                                                                 \\ \bottomrule
\end{tabular}
\label{tab:workers_statistics}
\end{table}
% \FloatBarrier

%% file: figs_and_tables/fig_spymasters_statistics.tex
\begin{figure}[!htb]
    \centering
    \subfloat[\centering 5 \& 6 Candidates]{{\includegraphics[width=5cm]{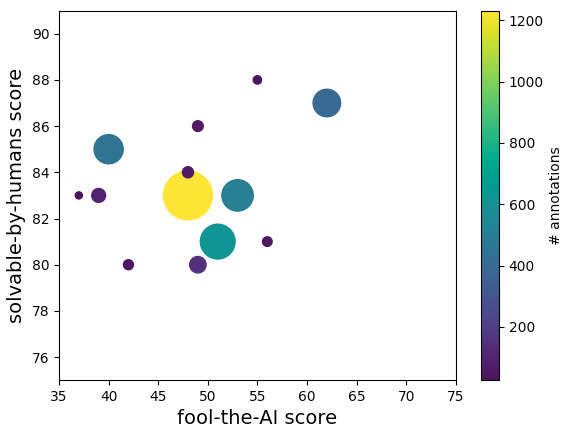} }}
    \qquad
    \subfloat[\centering 10 \& 12 Candidates]{{\includegraphics[width=5cm]{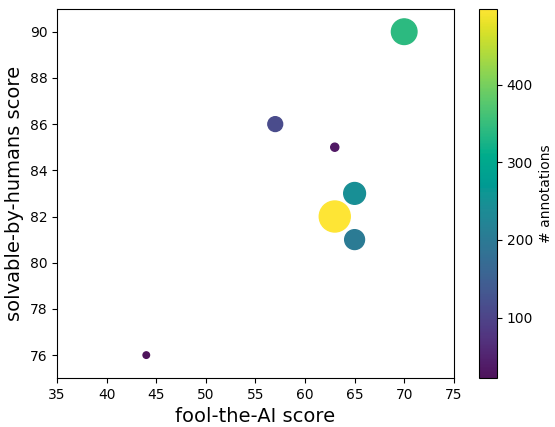} }}
    \caption{Spymasters fool-the-AI and solvable-by-human scores. Each point represents a spymaster. The best spymaster on the top right achieved fool-the-AI score of 62 and solvable-by-humans score of 87 on the case of 5 \& 6 candidates; and a fool-the-AI score of 70 and solvable-by-humans score of 90 on the case of 10 \& 12 candidates}
    \label{fig:spymasters_stats}
\end{figure}

%% file: figs_and_tables/fig_statistics.tex
\begin{figure}[!htb]
\centering
\newcommand{\figlen}[0]{\columnwidth}
    \includegraphics[width=0.5\figlen]{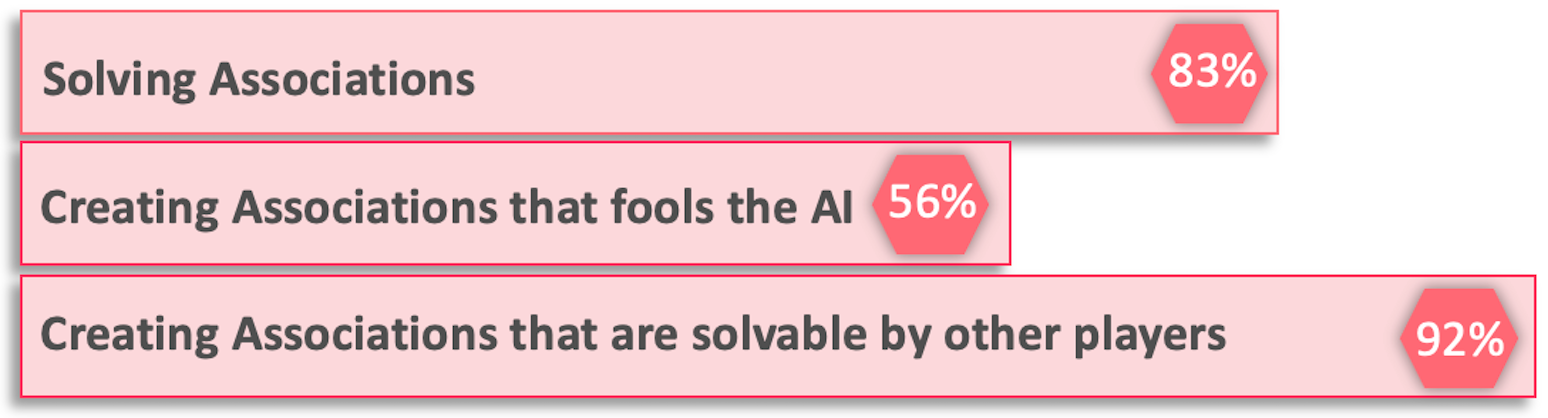}\\
    \caption{A screenshot from the player dashboard, aiming to increase players motivation. It contains different statistics measuring the performance in beating the AI, creating novel associations, and solving other player's associations.}
    \label{fig:player_statistics}
\end{figure}

%% file: figs_and_tables/fig_concreteness.tex
\begin{figure}[!htb]
\centering
\newcommand{\figlen}[0]{\columnwidth}
    \includegraphics[width=0.5\figlen]{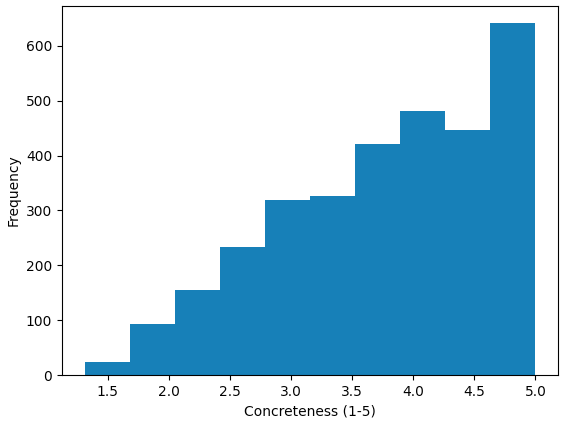}\\
     \caption{Generated cues concreteness distribution.}
     \label{fig:concreteness}
\end{figure}
% \FloatBarrier

%% file: figs_and_tables/table_human_performance_on_candidates.tex
Table~\ref{tab:human_performance_on_candidates} show results for all cases of generated data, with different number of candidates and generated associations. We observe that spymasters usually selected two associations, and that performance (both human and model) are similar between 5 and 6, and between 10 and 12. When comparing human to model performance, we see that the generated data is challenging for models and easy for humans.

\begin{table}[!htb]
\caption{\gamesplit{} Human and model (CLIP RN50) for different candidates and distractors}
\centering
\begin{tabular}{@{}lllll@{}}
\toprule
\# Candidates      & \# Associations ($k$) & \# Items & \% Human Performance & \% Model Performance \\ \midrule
\multirow{2}{*}{5} & 2               & 1,091      & 90  & 52        \\
                   & 3               & 234      & 92   & 57        \\ \midrule
\multirow{3}{*}{6} & 2               & 1,087      & 90   & 48        \\
                   & 3               & 259      & 88    & 51       \\
                   & 4               & 43       & 100     & 57      \\ \midrule
\multirow{4}{*}{10} 
& 2               & 338      & 87  & 37        \\
                   & 3               & 83      & 93   & 35        \\ 
& 4               & 5      & 92  & 29        \\
                   
                   \midrule
\multirow{4}{*}{12} & 2               & 328      & 90   & 37        \\
                   & 3               & 84      & 93    & 33       \\
                   & 4               & 16       & 100     & 28 \\
 \bottomrule

\end{tabular}
\label{tab:human_performance_on_candidates}
\end{table}
% \FloatBarrier

%% file: figs_and_tables/model_analysis_grid.tex
\begin{figure}[ht] % <---
  \begin{subfigure}{0.30\textwidth}
      \includegraphics[width=\linewidth]{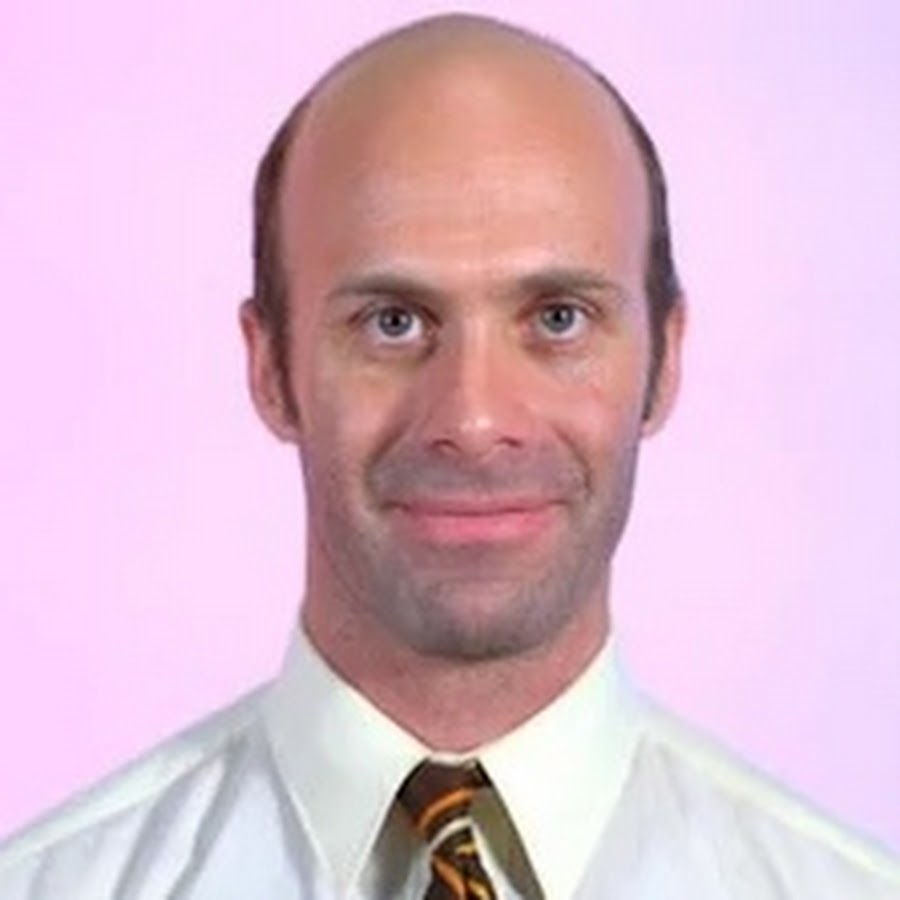}
      \caption{human}
  \end{subfigure}
\hfill % <--- 
  \begin{subfigure}{0.30\textwidth}
      \includegraphics[width=\linewidth]{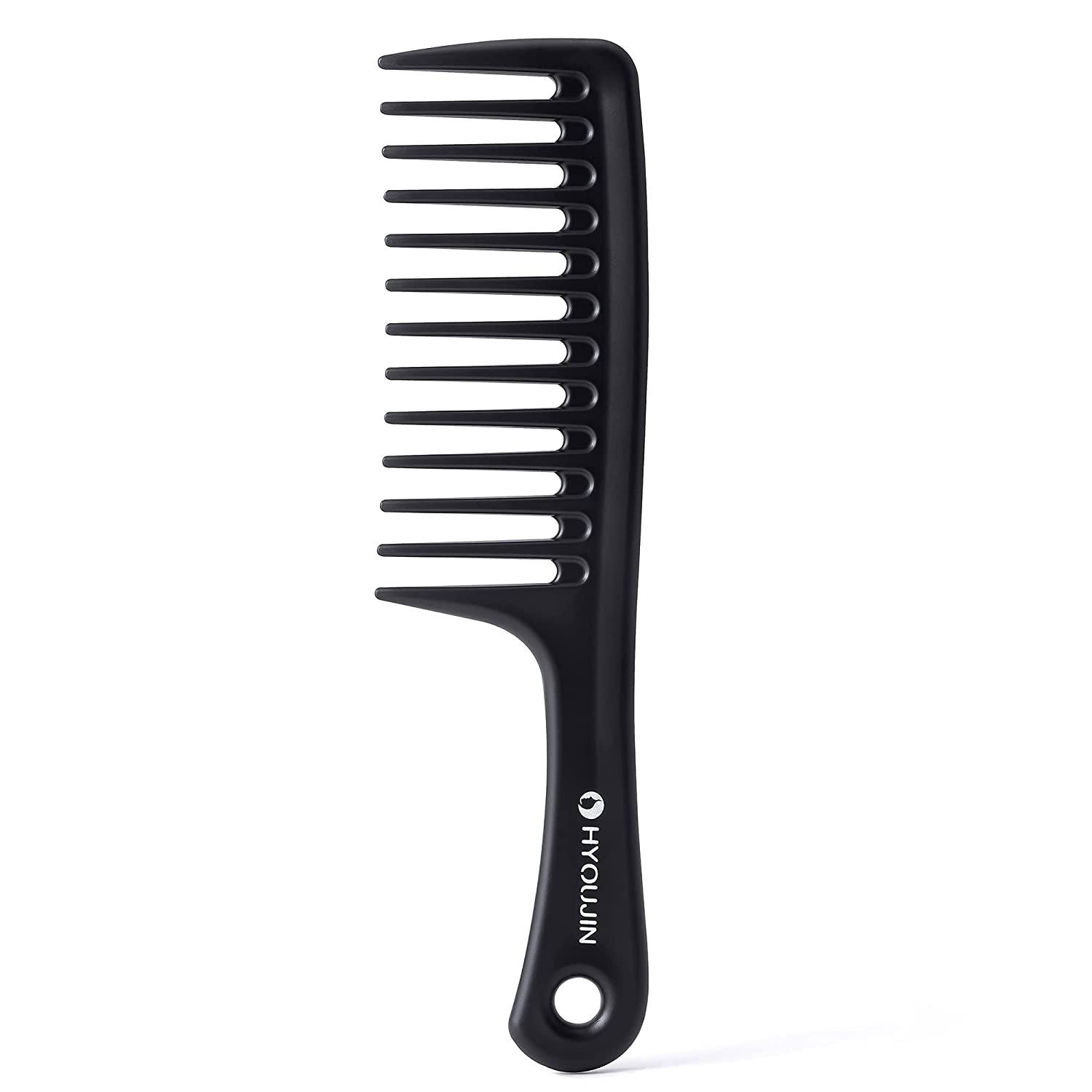}
      \caption{comb}
  \end{subfigure}
\hfill % <---
  \begin{subfigure}{0.30\textwidth}
      \includegraphics[width=\linewidth]{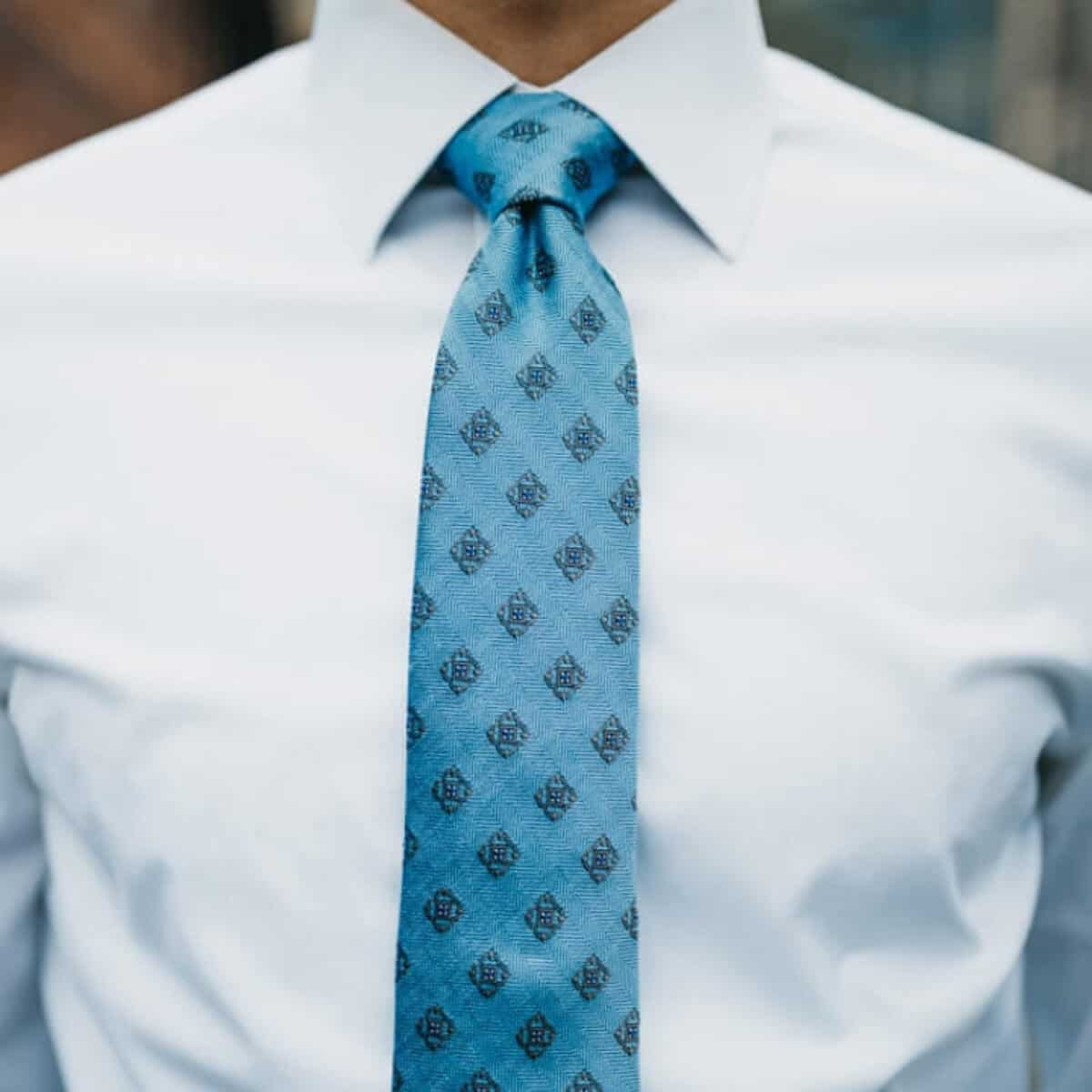}
      \caption{tie}
  \end{subfigure}

  \caption{Visually salient}
      \label{fig:visually_salient}

  \begin{subfigure}{0.30\textwidth}
      \includegraphics[width=\linewidth]{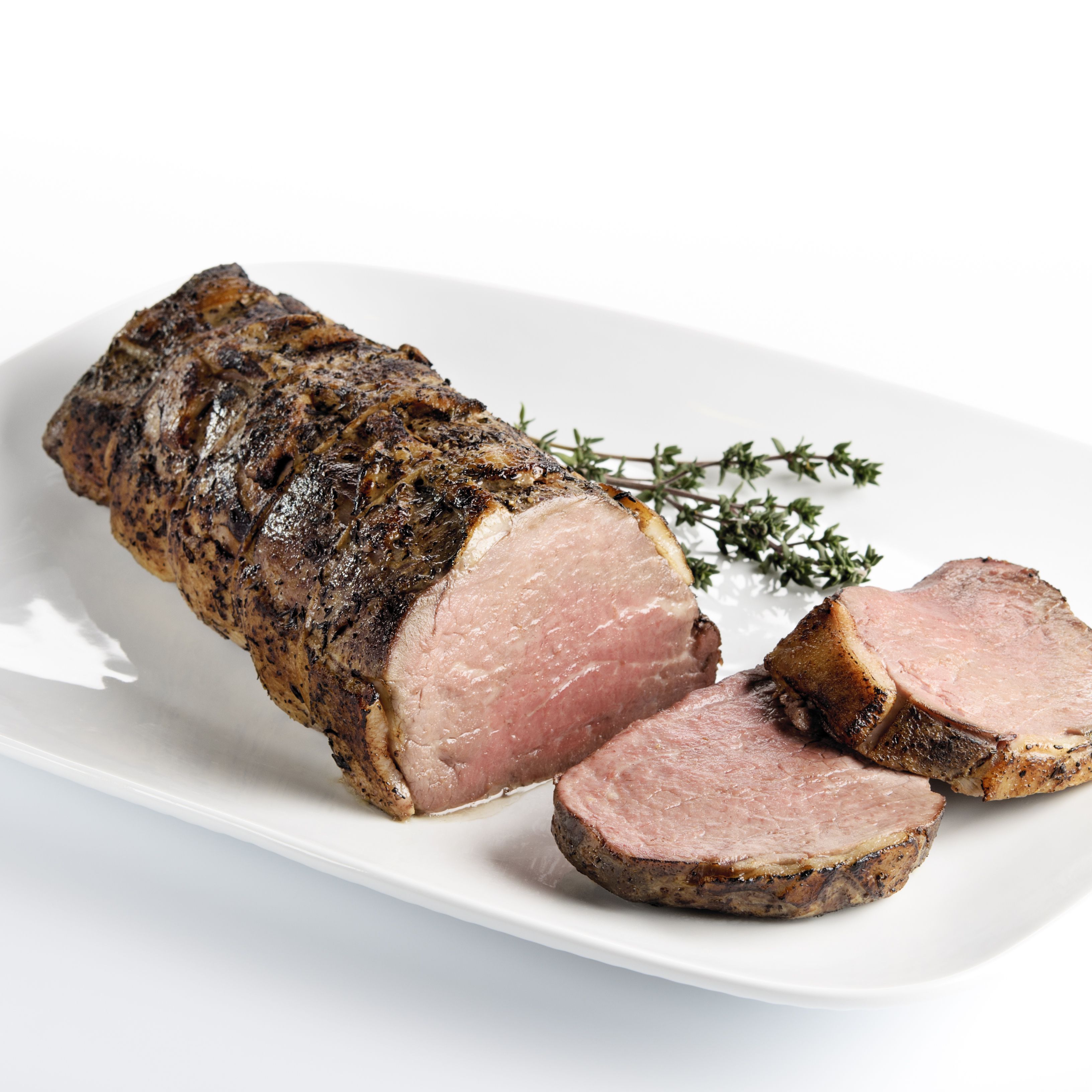}
      \caption{log}
  \end{subfigure}
\hfill % <--- 
  \begin{subfigure}{0.30\textwidth}
      \includegraphics[width=\linewidth]{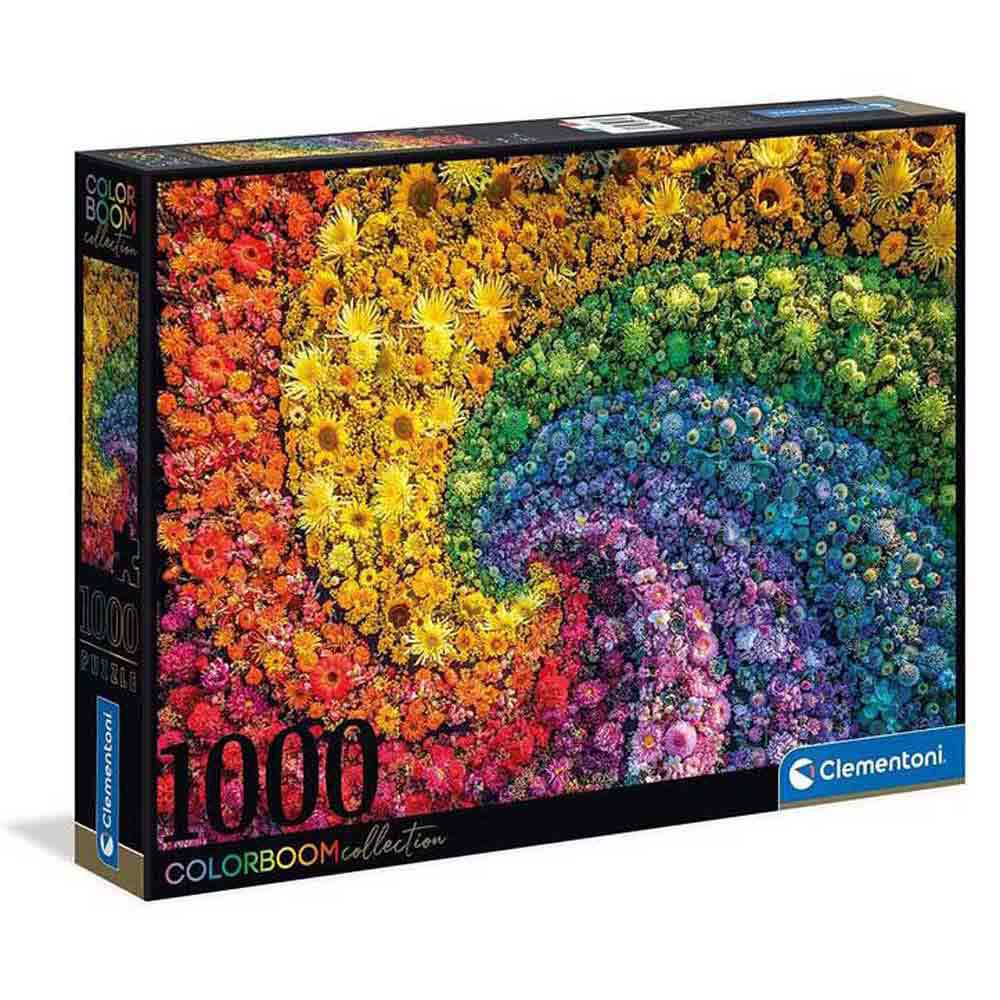}
      \caption{pride}
  \end{subfigure}
\hfill % <---
  \begin{subfigure}{0.30\textwidth}
      \includegraphics[width=\linewidth]{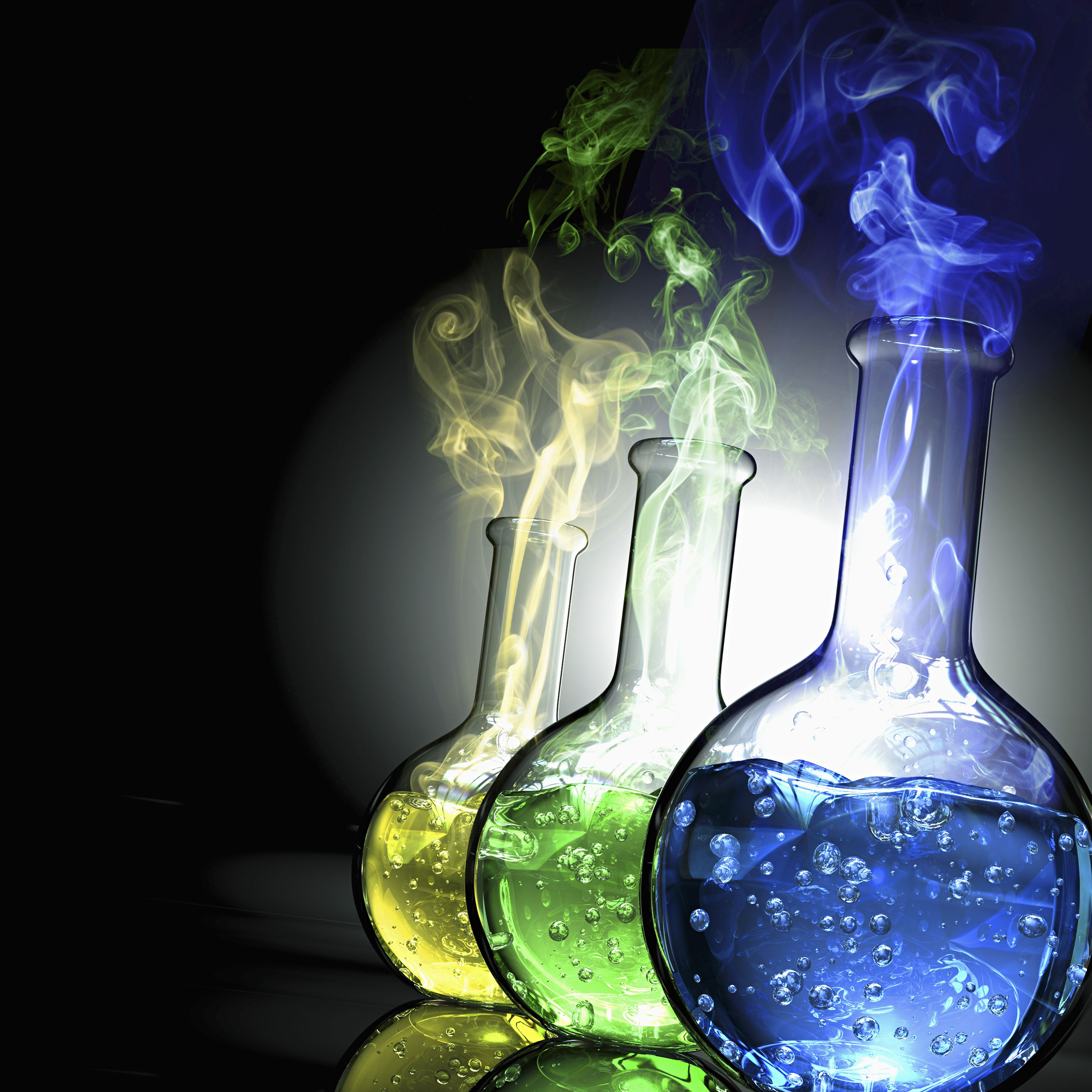}
      \caption{vapors}
  \end{subfigure}

  \caption{Visually non-salient}
     \label{fig:visually_non_salient}

  \begin{subfigure}{0.30\textwidth}
      \includegraphics[width=\linewidth]{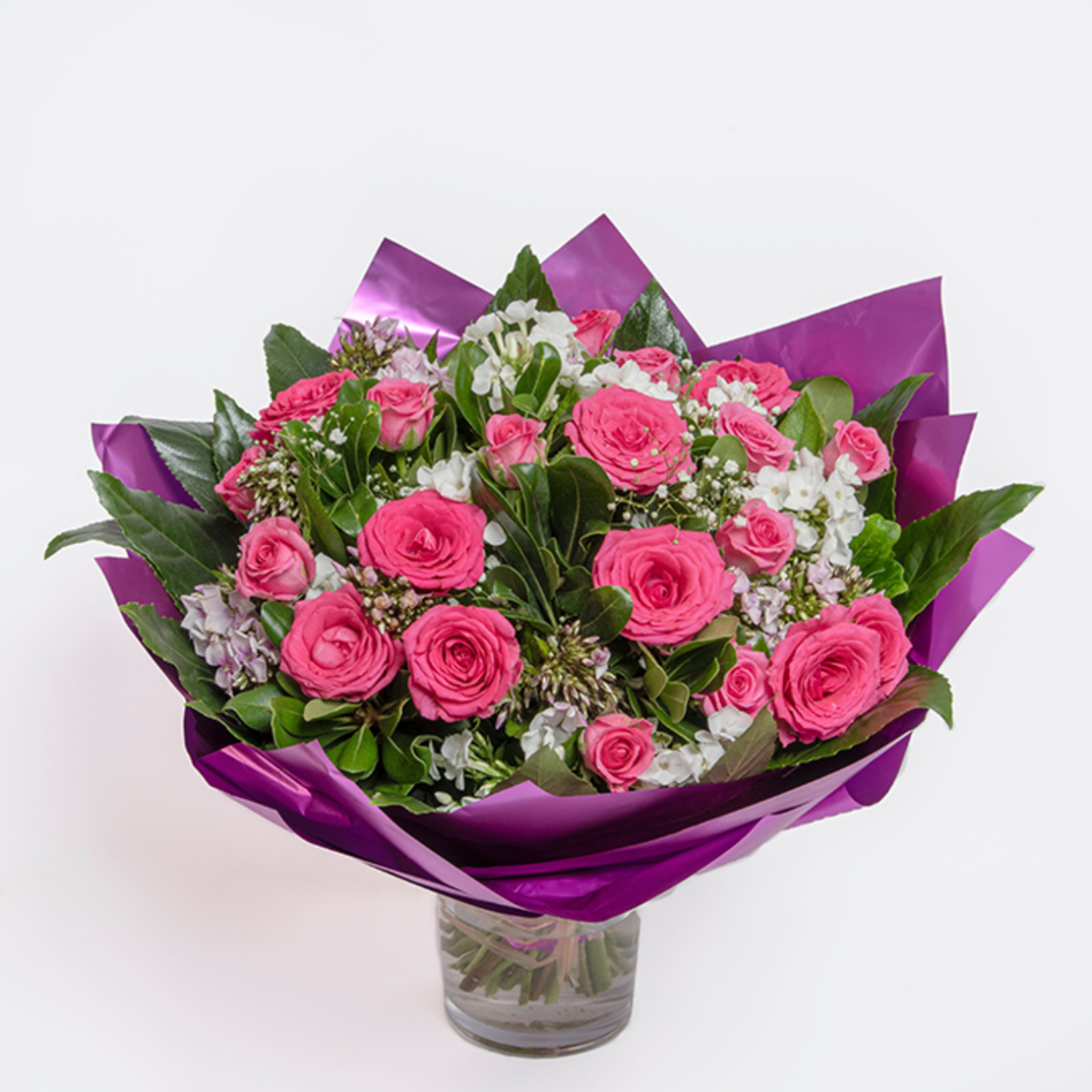}
        \caption{pollinate}
  \end{subfigure}
\hfill % <--- 
  \begin{subfigure}{0.30\textwidth}
      \includegraphics[width=\linewidth]{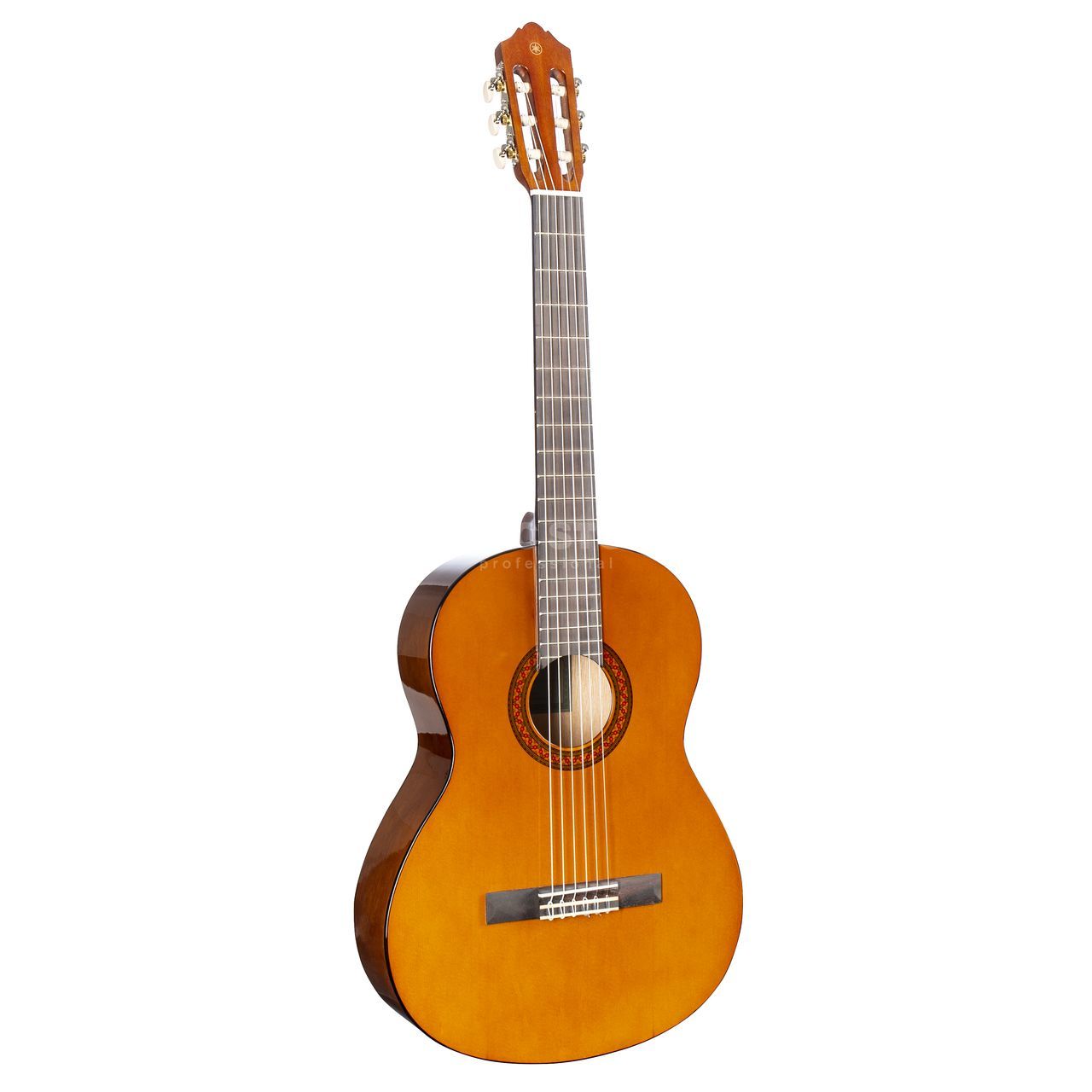}
        \caption{loud}
  \end{subfigure}
\hfill % <---
  \begin{subfigure}{0.30\textwidth}
      \includegraphics[width=\linewidth]{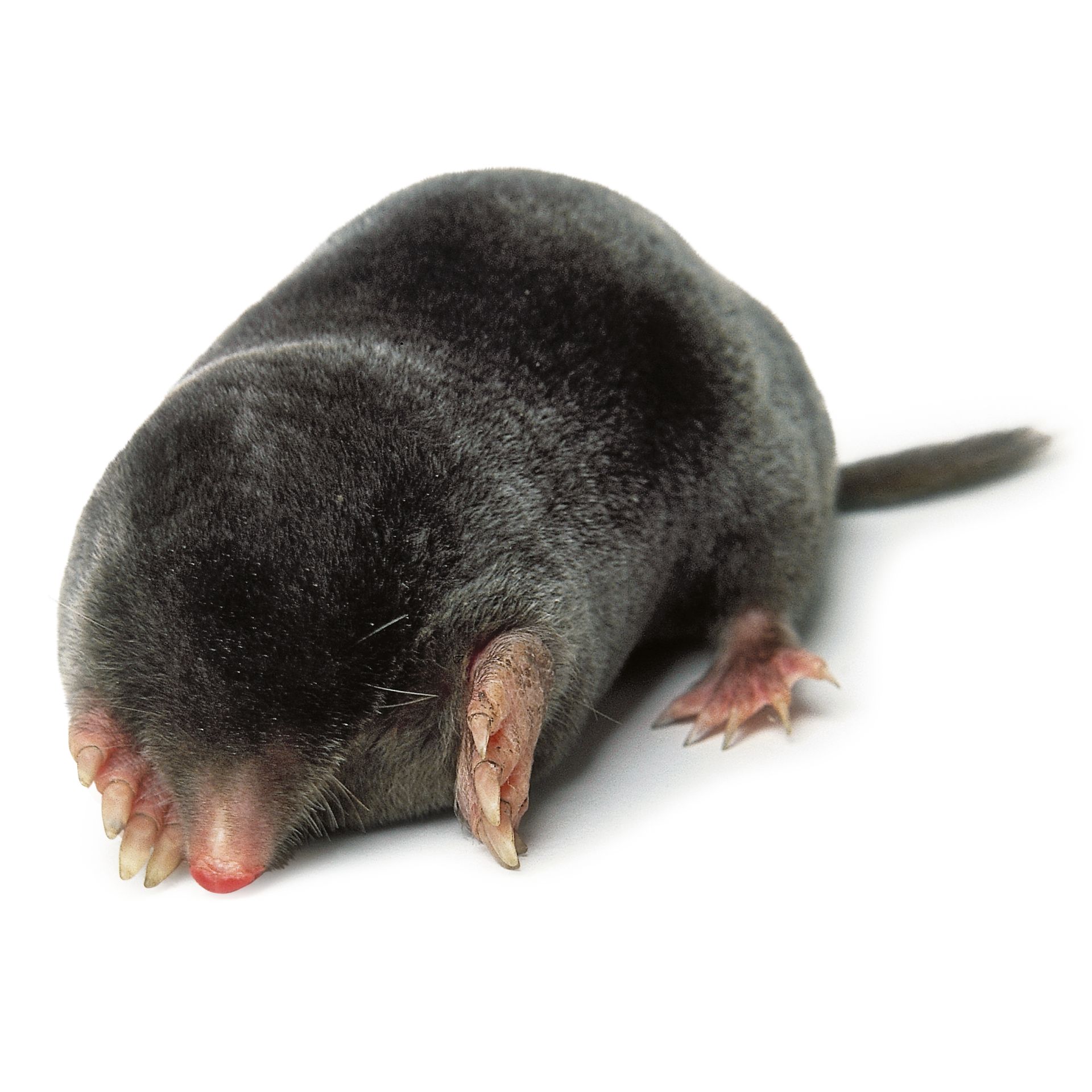}
        \caption{lawn}
  \end{subfigure}

  \caption{Concept related}
  \label{fig:grid_concept}
   
    \begin{subfigure}{0.30\textwidth}
      \includegraphics[width=\linewidth]{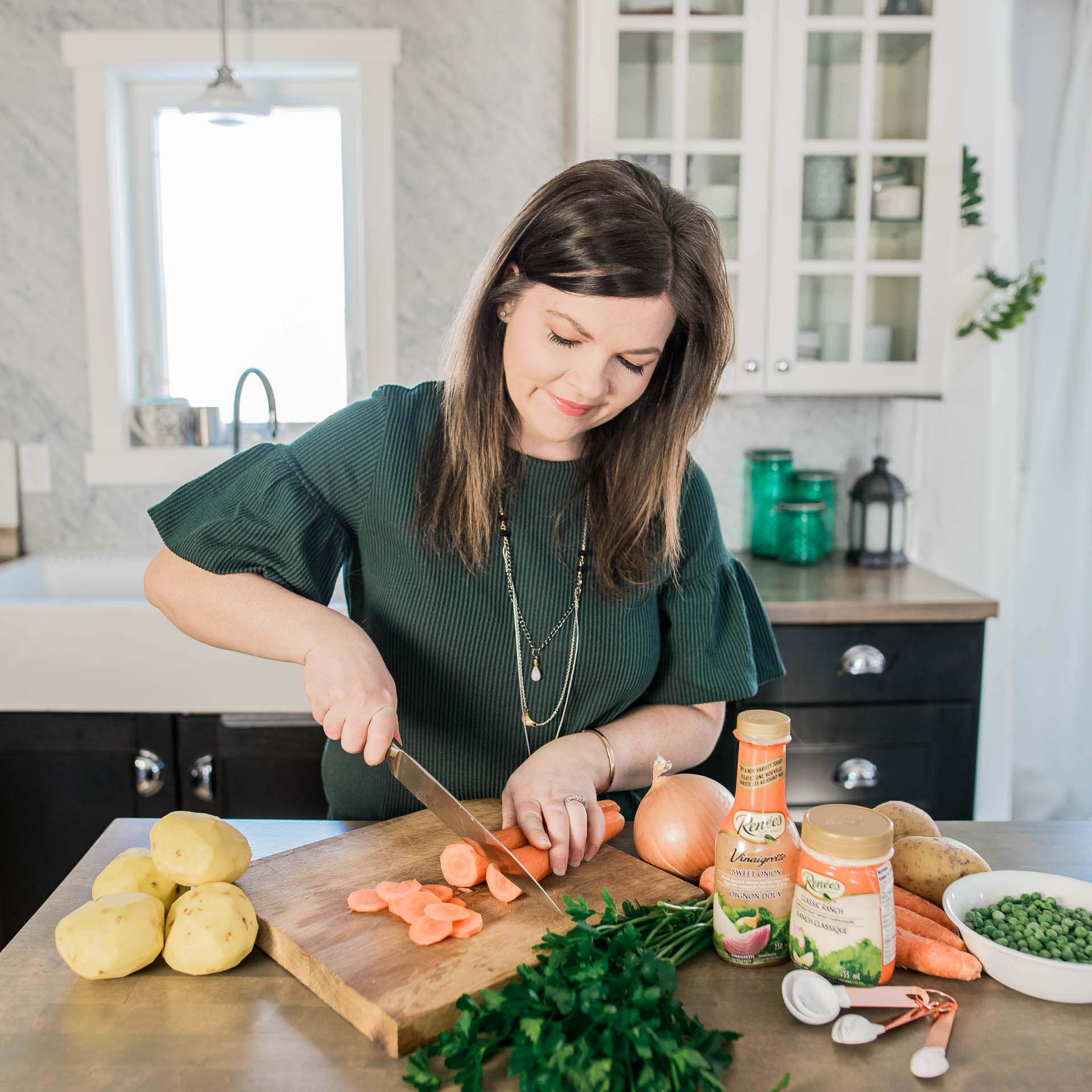}
      \caption{cook}
  \end{subfigure}
\hfill % <--- 
  \begin{subfigure}{0.30\textwidth}
      \includegraphics[width=\linewidth]{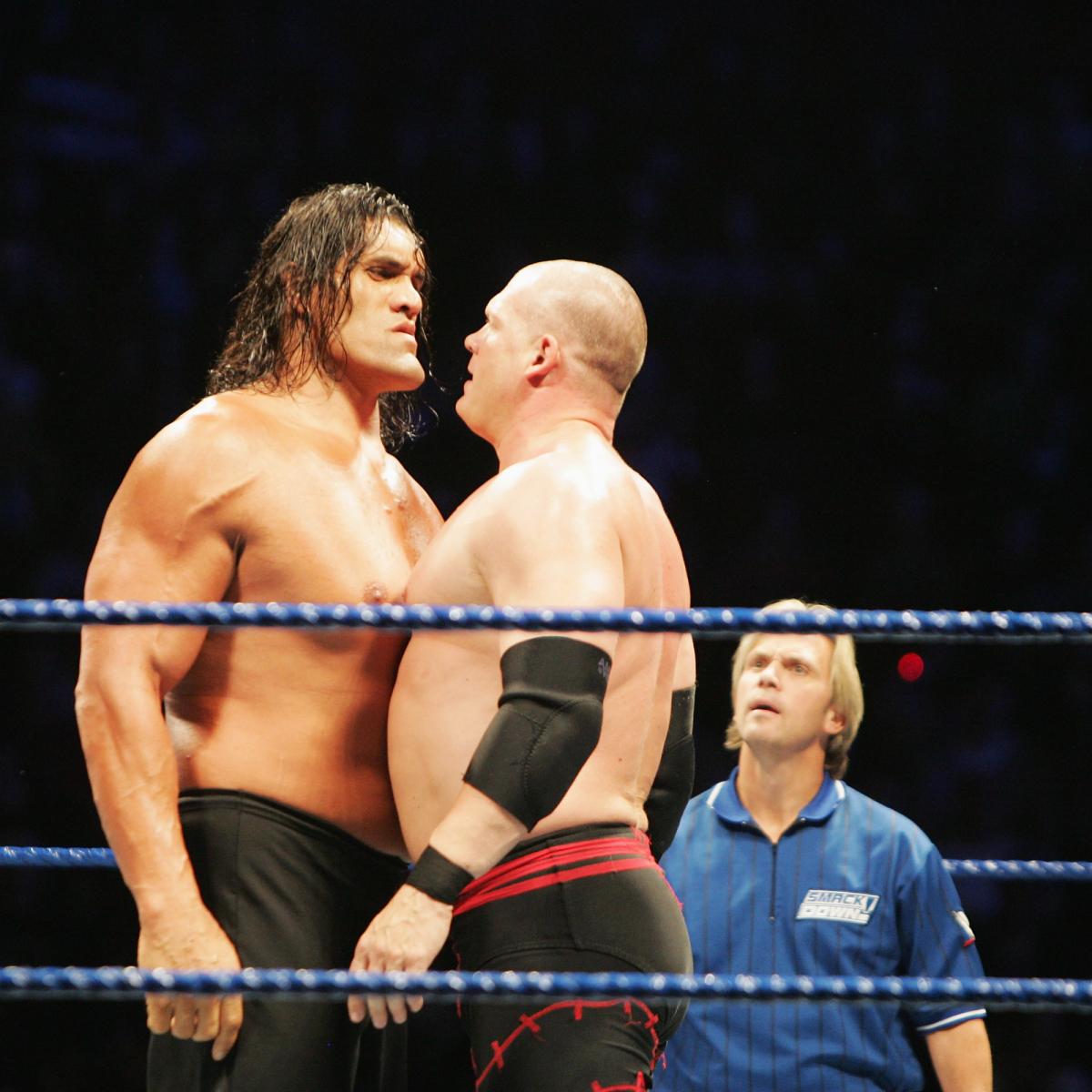}
      \caption{confront}
  \end{subfigure}
\hfill % <---
  \begin{subfigure}{0.30\textwidth}
      \includegraphics[width=\linewidth]{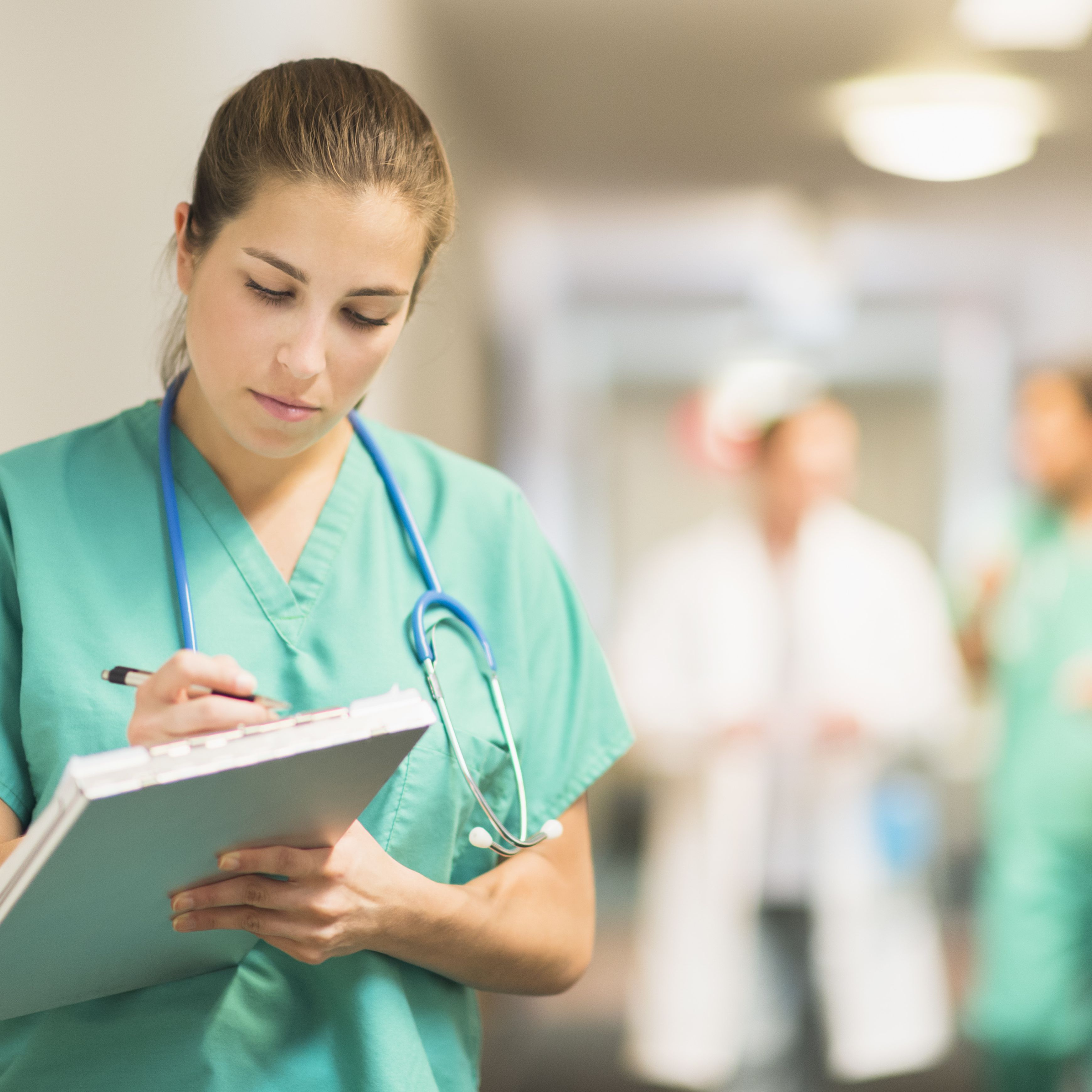}
      \caption{hold}
  \end{subfigure}

  \caption{Activity}
  \label{fig:grid_activity}
  \end{figure}

    \begin{figure}
     \begin{subfigure}{0.30\textwidth}
      \includegraphics[width=\linewidth]{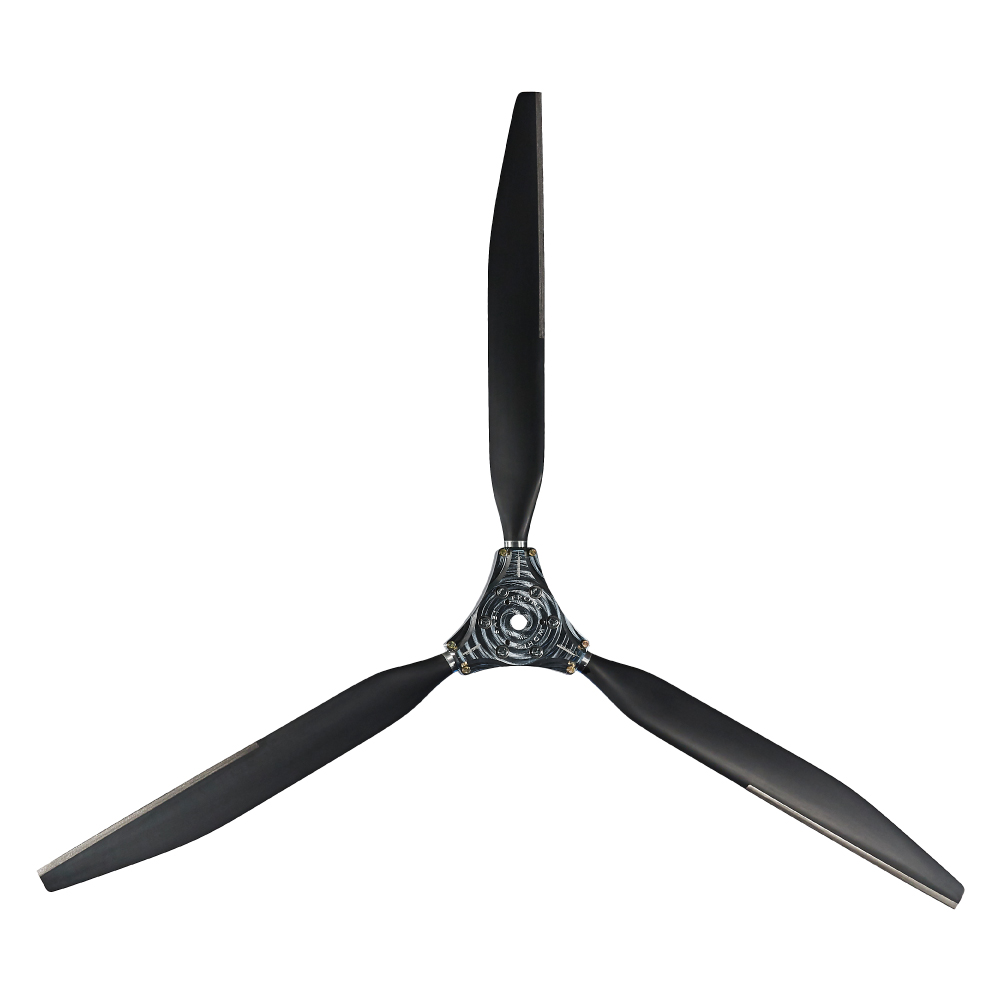}
      \caption{three}
  \end{subfigure}
\hfill % <--- 
  \begin{subfigure}{0.30\textwidth}
      \includegraphics[width=\linewidth]{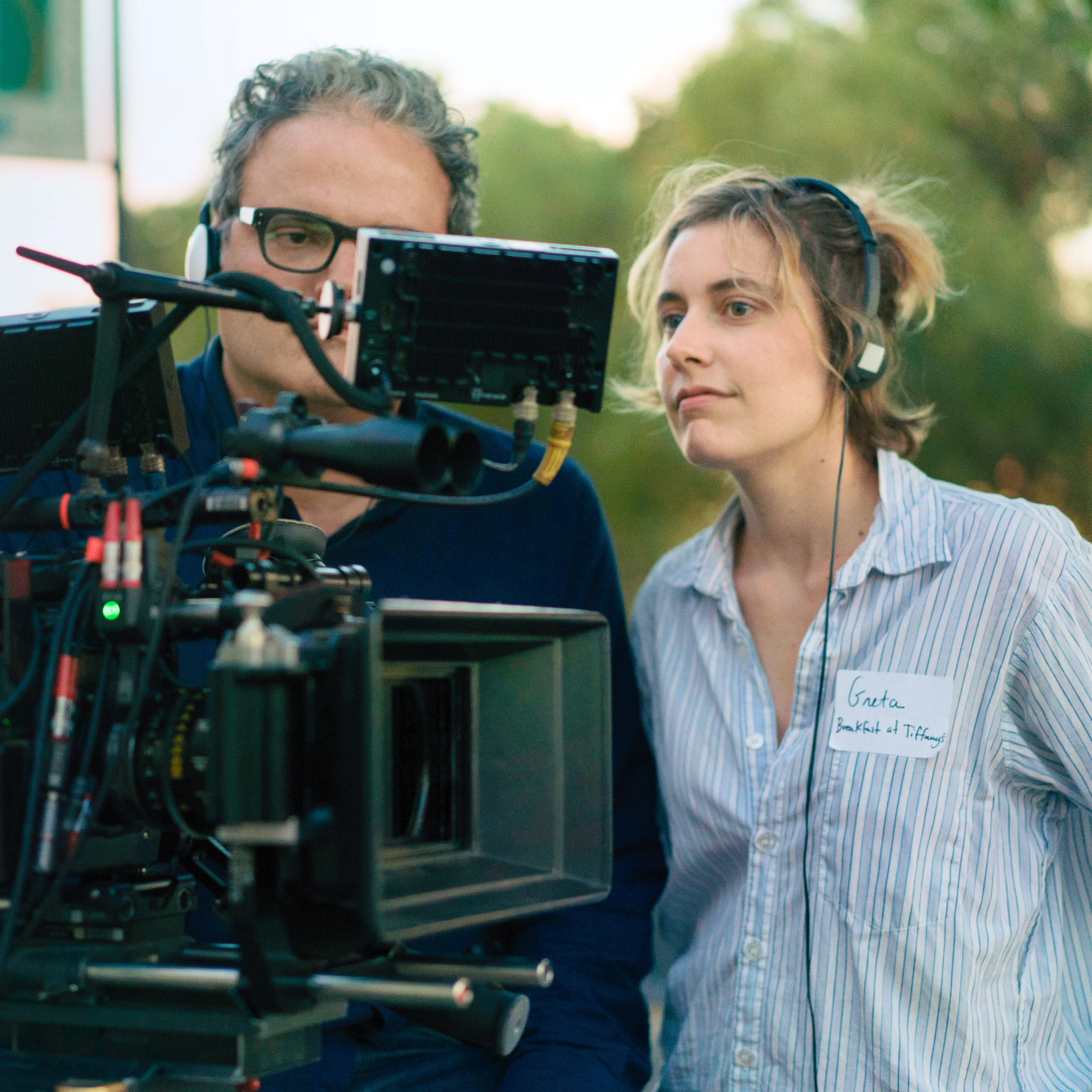}
      \caption{two}
  \end{subfigure}
\hfill % <---
  \begin{subfigure}{0.30\textwidth}
      \includegraphics[width=\linewidth]{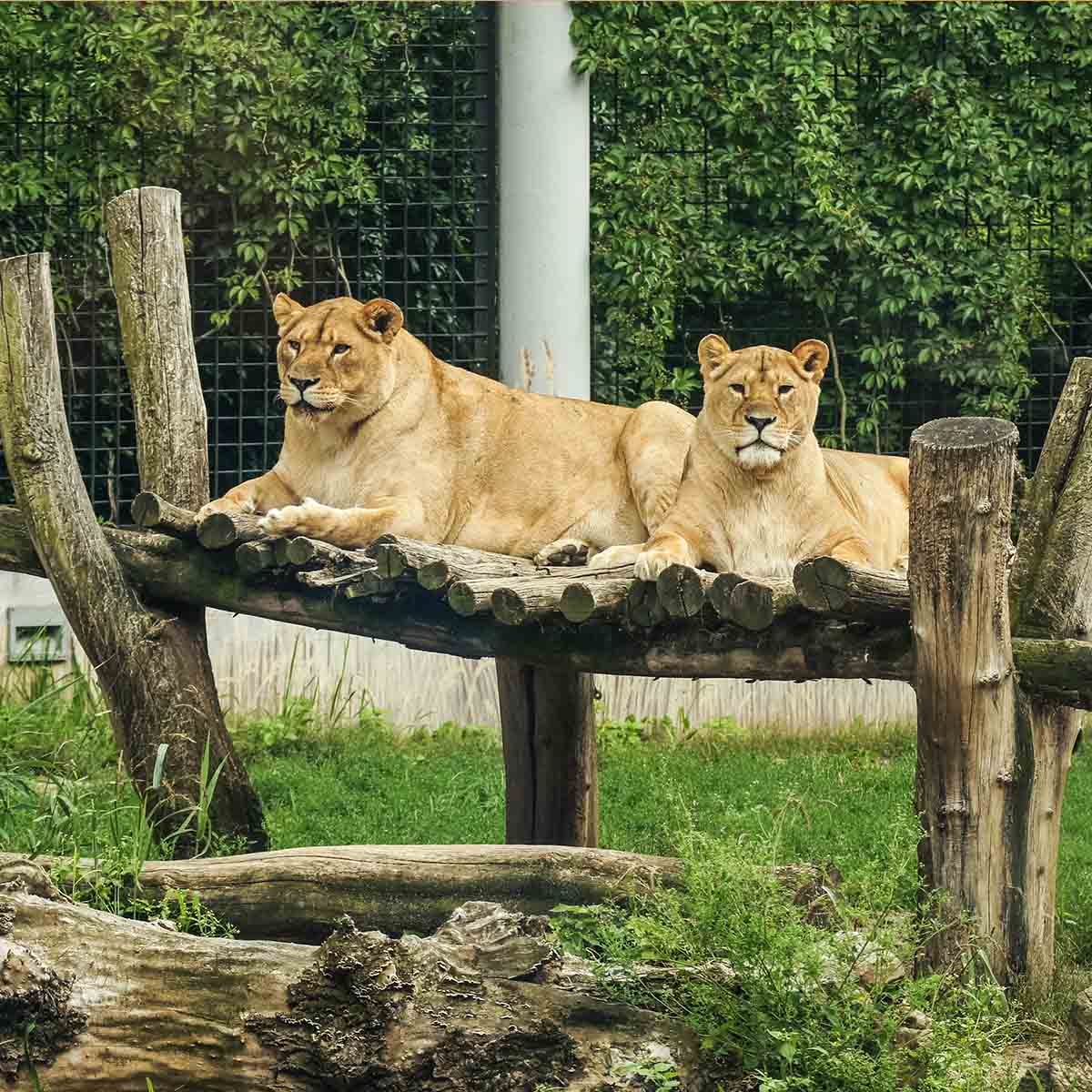}
      \caption{multiple}
  \end{subfigure}

  \caption{Aggregation / Counting}
  \label{fig:grid_counting}
  
   \begin{subfigure}{0.30\textwidth}
      \includegraphics[width=\linewidth]{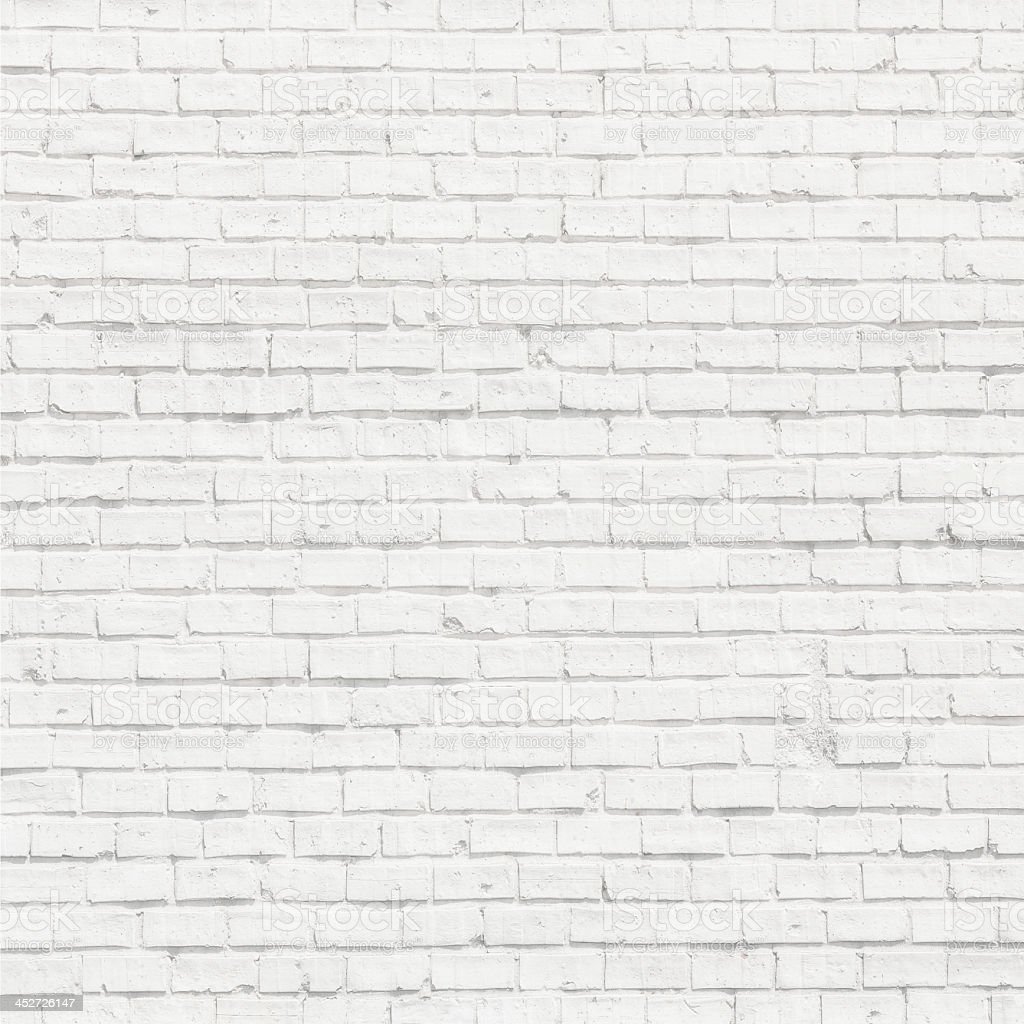}
      \caption{white}
  \end{subfigure}
\hfill % <--- 
  \begin{subfigure}{0.30\textwidth}
      \includegraphics[width=\linewidth]{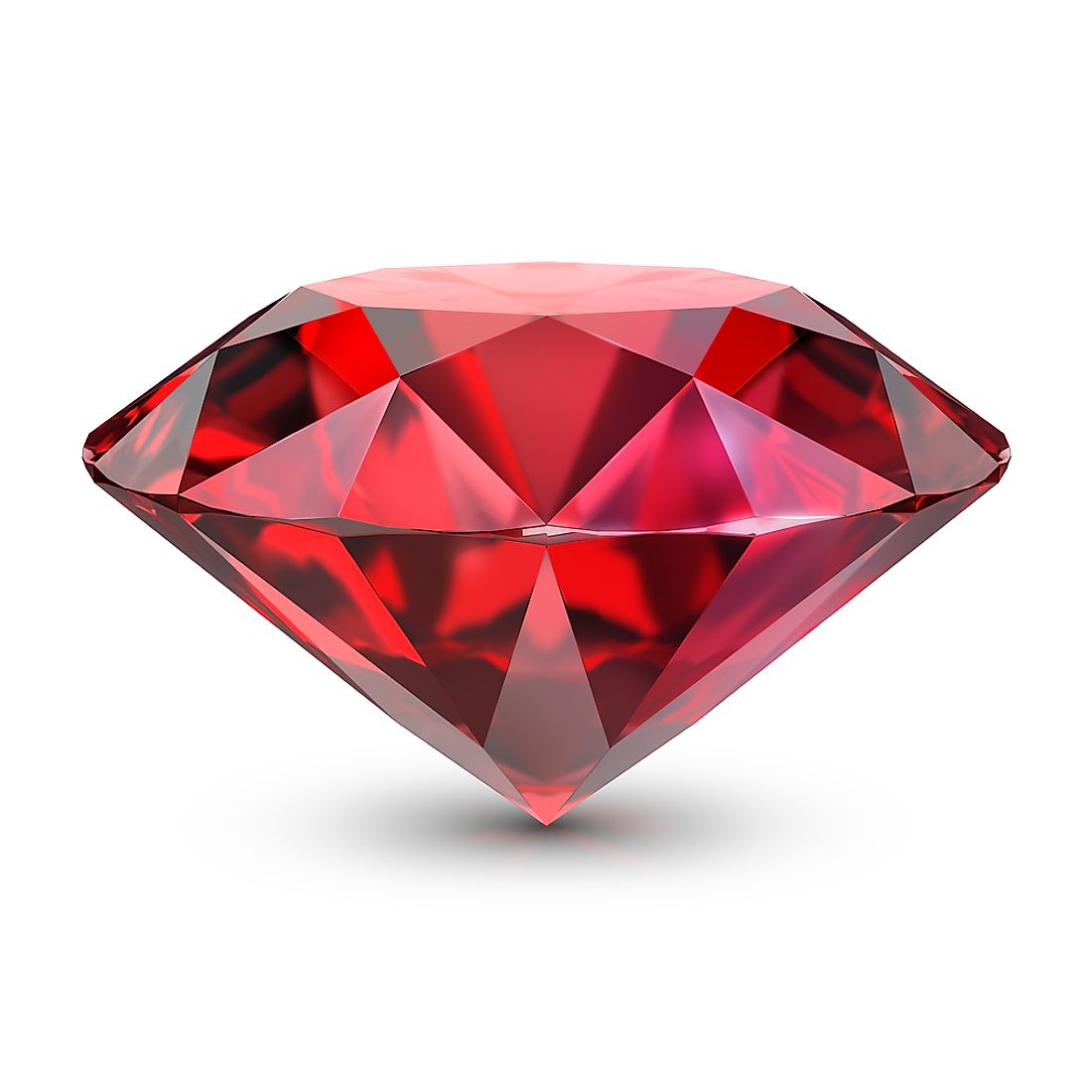}
      \caption{sanguine}
  \end{subfigure}
\hfill % <---
  \begin{subfigure}{0.30\textwidth}
      \includegraphics[width=\linewidth]{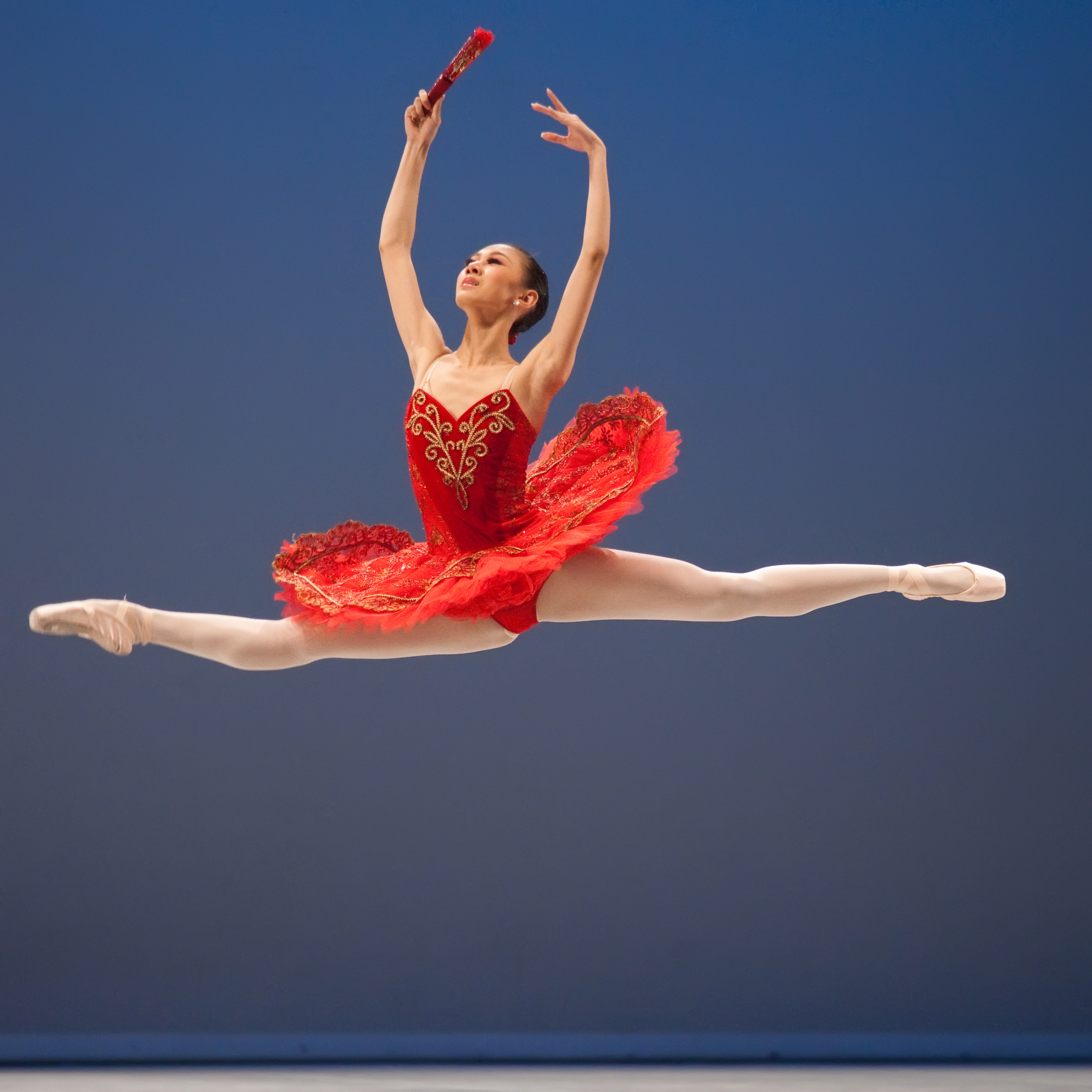}
      \caption{red}
  \end{subfigure}
  \caption{Colors}
  \label{fig:grid_colors}
  
\end{figure}

%% file: figs_and_tables/fig_example_mturk_association_type.tex
\begin{figure}[!htb]
\centering
\newcommand{\figlen}[0]{\columnwidth}
    \includegraphics[width=0.8\figlen]{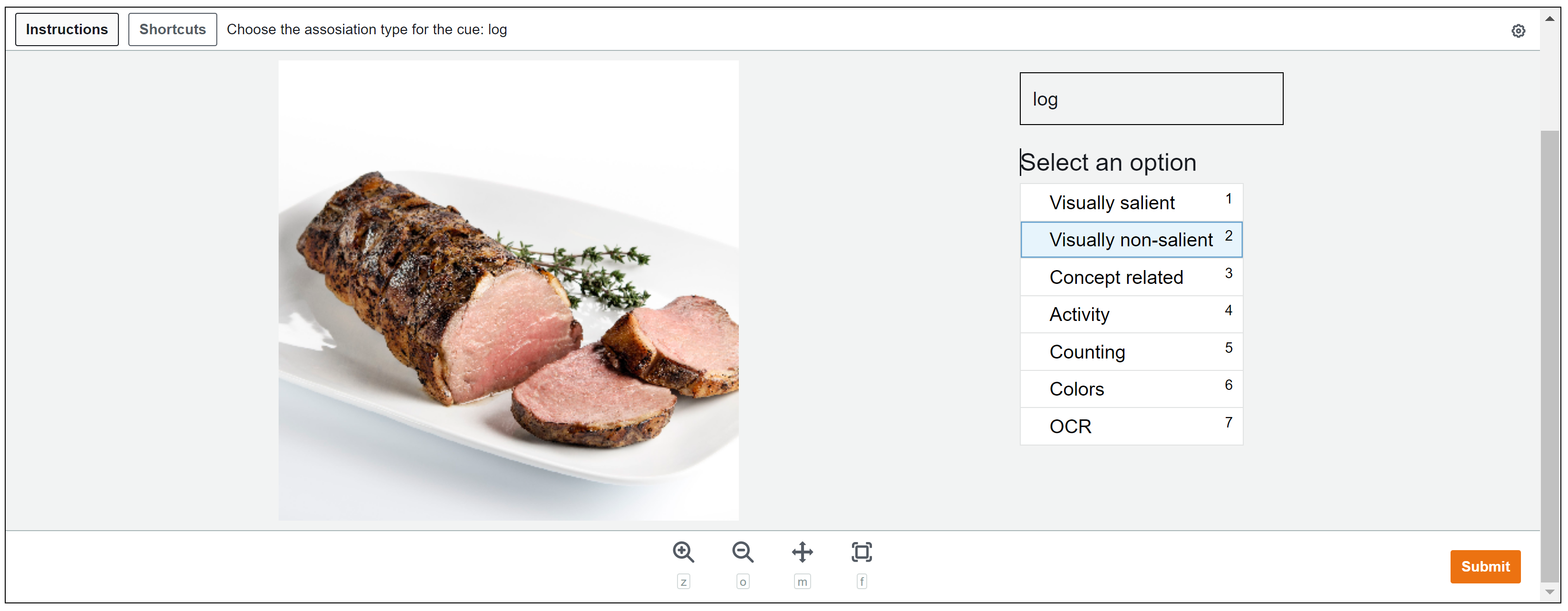}\\
     \caption{A screenshot from the task of annotating different analogy types.}
     \label{fig:mturk_association_type_task}
\end{figure}
% \FloatBarrier

%% file: sections/99A_swow_multimodal.tex
The \swowsplit{} has four options of text-image modalities, so we evaluate all cases of models: vision-and-language, textual only and visual only. 

\paragraph{Computer vision models} when both the cue and candidates are visual we evaluate ViT \cite{dosovitskiy2020image}, Swin Transformer \cite{liu2021swin}, DeiT \cite{touvron2021training} and ConvNeXt \cite{liu2022convnet}.\footnote{The exact versions we took are the largest pretrained versions available in \href{https://github.com/rwightman/pytorch-image-models}{timm} library: ViT Large patch32-384, Swin Large patch4 window7-224, DeiT Base patch16 384, ConvNeXt Large.} 

\paragraph{Visual associations are more difficult than textual}
Table~\ref{tab:table_swow_all_modalities} shows results for the different modalities. The performance is the highest in the all-text version, decreases when one of the cues or candidates are images, and the worst when both are images. 
\input{figs_and_tables/table_swow_all_modalities}

%% file: figs_and_tables/table_swow_all_modalities.tex
\begin{table}[!htb]
\centering
\caption{Results on the multi-modal versions of SWOW baseline dataset. Visual associations are more difficult than textual}
\label{tab:table_swow_all_modalities}
\begin{tabular}{@{}ccccc@{}} \toprule
\multirow{2}{*}{Model type}          & \multirow{2}{*}{Model} & \multicolumn{2}{c}{Modalities}             & \multirow{2}{*}{Jaccard Index} \\
                                     &                        & Key                        & Candidates    &                                \\ \midrule
\multirow{10}{*}{Vision and Language} & \multirow{4}{*}{CLIP-ViT-L/14} & \multirow{2}{*}{Text}      & Text          & 86                             \\
                                      &                            &                            & Image         & 74                             \\
                                      &                            & \multirow{2}{*}{Image}     & Text          & 79                             \\
                                      &                            &                            & Image         & 65                             \\
                                      & \multirow{2}{*}{ViLT}      & Text                       & Image         & 58                             \\
                                      &                            & Image                      & Text          & 59                             \\
                                      &   \multirow{2}{*}{LiT}     & Text                       & Image         & 37                              \\
                                      &                            & Image                      & Text          & 40                              \\ 
                                      & \multirow{2}{*}{X-VLM}     & Text                       & Image         & 68                              \\
                                      &                            & Image                      & Text          & 70                              \\ \midrule
\multirow{4}{*}{Vision}              & ViT                    & \multicolumn{2}{c}{\multirow{4}{*}{Image}} & 61                             \\
                                     & Swin                   & \multicolumn{2}{c}{}                       & 59                             \\
                                     & DeiT                   & \multicolumn{2}{c}{}                       & 53                             \\
                                     & ConvNeXt               & \multicolumn{2}{c}{}                       & 56                             \\ \midrule
\multirow{3}{*}{Text   Transformers} & MPNet                  & \multicolumn{2}{c}{\multirow{4}{*}{Text}}  & 88                             \\
                                     & MPNet QA               & \multicolumn{2}{c}{}                       & 91                             \\
                                     & Distil RoBERTa         & \multicolumn{2}{c}{}                       & 77                             \\ \midrule
Text Word2Vec                        & Spacy                  & \multicolumn{2}{c}{Text}                       & 91                 \\ \bottomrule 
\multirow{16}{*}{Text}  & CLIP-ViT-L/14                   & \multicolumn{2}{c}{\multirow{4}{*}{Text}}                             & 87                             \\
  & MPNet                  & \multicolumn{2}{c}{}                                                  & 88                             \\
  &  MPNet QA               & \multicolumn{2}{c}{}                                                  & 90                             \\
  & Distil RoBERTa         & \multicolumn{2}{c}{}                                                  & 73                             \\ \cmidrule(l){2-5}
  & CLIP-ViT-L/14                    & \multirow{4}{*}{Text}             & \multirow{4}{*}{Synthesized Text} & 55                             \\
  & MPNet                  &                                   &                                   & 72                             \\
  & MPNet QA               &                                   &                                   & 76                             \\
  & Distil RoBERTa         &                                   &                                   & 66                             \\ \cmidrule(l){2-5}
  & CLIP-ViT-L/14                   & \multirow{4}{*}{Synthesized Text} & \multirow{4}{*}{Text}             & 81                             \\
  & MPNet                  &                                   &                                   & 77                             \\
  & MPNet QA               &                                   &                                   & 78                             \\
  & Distil RoBERTa         &                                   &                                   & 73                             \\ \cmidrule(l){2-5}
  & CLIP-ViT-L/14                   & \multicolumn{2}{c}{\multirow{4}{*}{Synthesized Text}}                 & 61                             \\
  & MPNet                  & \multicolumn{2}{c}{}                                                  & 64                             \\
  & MPNet QA               & \multicolumn{2}{c}{}                                                  & 64                             \\
  & Distil RoBERTa         & \multicolumn{2}{c}{}                                                  & 67                            \\ \bottomrule
\end{tabular}\end{table}